\documentclass[conference]{IEEEtran}
\IEEEoverridecommandlockouts
\usepackage{cite}
\usepackage{amsmath,amssymb,amsfonts}
\usepackage{algorithmic}
\usepackage{graphicx}
\usepackage{textcomp}
\usepackage{xcolor}
\def\BibTeX{{\rm B\kern-.05em{\sc i\kern-.025em b}\kern-.08em
    T\kern-.1667em\lower.7ex\hbox{E}\kern-.125emX}}

\usepackage{array}
\usepackage{multirow}
\usepackage{color}
\usepackage{colortbl}
\usepackage{framed}
\usepackage{bm}
\usepackage{bbm}
\usepackage{xspace}
\usepackage{enumitem}
\usepackage{algorithm}
\usepackage{listings}
\usepackage{url}
\usepackage{booktabs}
\usepackage{pifont} 
\usepackage{comment}

\def \ie {\emph{i.e.}}
\def \eg {\emph{e.g.}}
\def \etal {\emph{et al.}}

\definecolor{LightCyan}{rgb}{0.88,0.95,1}
\definecolor{blond}{rgb}{0.98, 0.94, 0.75}
\definecolor{ourcolor}{gray}{0.91}

\newcommand{\cmark}{\ding{51}}%
\newcommand{\xmark}{\ding{55}}%

\newcommand{\ours}{Fashion-RAG\xspace}

\newcommand{\tit}[1]{\smallbreak\noindent\textbf{#1.}}
\newcommand{\tinytit}[1]{\noindent\textbf{#1.}}

\begin{document}

\title{Fashion-RAG: Multimodal Fashion Image Editing\\via Retrieval-Augmented Generation}


\author{\IEEEauthorblockN{Fulvio Sanguigni\textsuperscript{1,2}, Davide Morelli\textsuperscript{1,2}, Marcella Cornia\textsuperscript{1}, Rita Cucchiara\textsuperscript{1}}
\IEEEauthorblockA{\textsuperscript{1}\textit{University of Modena and Reggio Emilia, Italy}\\
\textsuperscript{2}\textit{University of Pisa, Italy}}
\texttt{\small\{name.surname\}@unimore.it}
}

\maketitle

\begin{abstract}
In recent years, the fashion industry has increasingly adopted AI technologies to enhance customer experience, driven by the proliferation of e-commerce platforms and virtual applications. Among the various tasks, virtual try-on and multimodal fashion image editing -- which utilizes diverse input modalities such as text, garment sketches, and body poses -- have become a key area of research. Diffusion models have emerged as a leading approach for such generative tasks, offering superior image quality and diversity. However, most existing virtual try-on methods rely on having a specific garment input, which is often impractical in real-world scenarios where users may only provide textual specifications. To address this limitation, in this work we introduce Fashion Retrieval-Augmented Generation (\ours), a novel method that enables the customization of fashion items based on user preferences provided in textual form. Our approach retrieves multiple garments that match the input specifications and generates a personalized image by incorporating attributes from the retrieved items. To achieve this, we employ textual inversion techniques, where retrieved garment images are projected into the textual embedding space of the Stable Diffusion text encoder, allowing seamless integration of retrieved elements into the generative process. Experimental results on the Dress Code dataset demonstrate that \ours outperforms existing methods both qualitatively and quantitatively, effectively capturing fine-grained visual details from retrieved garments. To the best of our knowledge, this is the first work to introduce a retrieval-augmented generation approach specifically tailored for multimodal fashion image editing.
\end{abstract}


\section{Introduction}
\label{sec:intro}
The rapid growth of e-commerce and online retail platforms has driven an increasing demand for enhanced user experiences in virtual environments. In the fashion domain, personalization and interactivity are crucial factors influencing customer satisfaction. Consequently, Computer Vision and Deep Learning solutions have emerged as key enablers, with generative models playing a pivotal role in addressing fashion-specific tasks. Early solutions in this domain primarily relied on Generative Adversarial Networks (GANs)~\cite{goodfellow2014generative}. Following this research path, several works leveraged GAN-based solutions to tackle fashion-oriented tasks~\cite{han2018viton,choi2021viton,fincato2021viton,morelli2022dresscode}, demonstrating promising results in generating plausible images. However, GANs suffer from mode collapse, limited fine-grained details, and spatial structure preservation ability. More recently, diffusion models~\cite{sohl2015deep,ho2020denoising} have emerged as a better alternative, offering superior image quality and diversity by iteratively refining noise-added images. Latent Diffusion Models (LDMs)~\cite{rombach2022high,esser2024scaling} have been proposed to reduce the computational cost associated with early diffusion models, significantly improving efficiency while maintaining high-quality visual outputs.

Expanding on these advancements, diffusion models have unlocked new frontiers in fashion-related research, including virtual try-on~\cite{morelli2023ladi,kim2024stableviton,wang2024stablegarment,choi2024idmvton} and multimodal fashion image editing~\cite{baldrati2023multimodal,baldrati2024multimodal,wang2024texfit}. While virtual try-on models can generally lead to better results, they require the try-on garment as input, which may not always be available in practical scenarios. Instead, multimodal fashion editing approaches leverage multimodal inputs like text, garment sketches, and body poses to generate a new fashion item directly worn by a model. Given the lack of a garment input image as a constraint for the generative architecture, these approaches can struggle with rendering fine-grained visual details of complex garments and correctly synthesizing visual attributes such as color, texture, and patterns. These shortcomings become more pronounced when dealing with challenging scenarios, such as the generation of intricate in-shop garments or models in dynamic poses.

Motivated by these challenges, we draw inspiration from the NLP field and introduce a novel Retrieval-Augmented Generation (RAG) approach for the visual domain. In NLP, RAG techniques~\cite{gao2023retrieval} have been adopted to address hallucination issues in text-only 
and multimodal LLMs, where large-scale architectures may generate inaccurate information. By incorporating additional knowledge retrieved from external sources during the generative process, RAG frameworks~\cite{asai2023selfrag,rao2024raven,gao2023retrieval} improve the factual consistency and relevance of generated content. We hypothesize that similar principles can be applied to fashion-oriented multimodal image synthesis, where retrieving relevant garments can enhance both the realism and fidelity of fashion-based generative results.

Following this intuition, we propose \ours, a RAG-based framework tailored specifically for multimodal fashion image editing. Given a user-provided garment textual description, the proposed method uses it as a query to retrieve multiple garments and integrates their attributes into the generative process. To facilitate seamless incorporation of retrieved elements, we employ a textual inversion technique~\cite{gal2022textual,baldrati2023zeroshot,morelli2023ladi} that projects garment images into the textual embedding space of CLIP~\cite{Radford2021LearningTV}, commonly used by text-to-image diffusion models~\cite{rombach2022high,podell2023sdxl}. This enables our model to effectively blend fine-grained details from retrieved garments with the user’s textual input, resulting in high-quality personalized outputs. Extensive experiments on the Dress Code dataset~\cite{morelli2022dresscode}, which contains multimodal annotations of multiple garment categories (\ie, upper-body, lower-body, and full-body clothes), demonstrate the benefits of incorporating retrieved garments during the generation process and show that \ours can achieve better results in terms of realism and coherence with the given inputs compared to state-of-the-art approaches. 

\smallskip
\noindent \textbf{Contributions. }To sum up, our contributions are as follows:
\begin{itemize}[left=3mm,noitemsep,topsep=0pt]
    \item We introduce a novel task that leverages RAG-based techniques to improve the quality and fidelity of diffusion-based multimodal fashion generation.
    \item We design a RAG framework, \ours, that incorporates fine-grained attributes from multiple retrieved garments using textual inversion techniques, enabling more accurate and visually consistent image synthesis.
    \item Comprehensive experiments conducted on the Dress Code dataset demonstrate that our approach outperforms existing methods both qualitatively and quantitatively, showing the effectiveness of incorporating retrieved garments to improve fashion-oriented generative tasks.
\end{itemize}

\section{Related Work}
\label{sec:related}

\tinytit{Diffusion-based Generative Models for Fashion} 
Diffusion models~\cite{dhariwal2021diffusion} have gained significant attention in fashion-related applications, consistently outperforming GANs in terms of image quality and alignment with control inputs. In particular, Latent Diffusion Models (LDMs)~\cite{rombach2022high,podell2023sdxl} are usually preferred, given their ability to reduce the computational load while preserving the diffusion generation capabilities. These models can be conditioned on fashion-oriented information through distinct methodologies. In the domain of virtual try-on, where the objective is to generate a realistic image of a person wearing a specified garment, two primary approaches have emerged. One approach follows a two-stage pipeline, in which the garment is first warped to fit the target model pose, and then the modified garment serves as the input of a generation network~\cite{gou2023taming,morelli2023ladi,chen2023size}. In contrast, other works~\cite{zhu2023tryondiffusion,kim2024stableviton,choi2024idmvton,chong2025catvton} employ a unified one-pass framework, where warping and generation occur simultaneously by coupling an auxiliary network with the primary generative component -- typically the Stable Diffusion U-Net~\cite{ronneberger2015u} -- to exploit contextual information learned during training. 

Among these, StableVITON~\cite{kim2024stableviton} and TryOnDiffusion~\cite{zhu2023tryondiffusion} introduce a learnable encoder copy of the original U-Net to extract garment features, which are then provided as inputs to the cross-attention layers of the original U-Net decoder. IDM-VTON~\cite{choi2024idmvton} extends this design by encoding the garment with two components: a full encoder-decoder U-Net and IP-Adapter~\cite{ye2023ip}. The features extracted from these components are utilized in the self-attention and cross-attention layers of the main U-Net. While these one-stage approaches demonstrate compelling results, they often suffer from garment-image misalignment issues. DCI-VTON~\cite{gou2023taming} addresses this misalignment by using both a spatially warped garment as an input to the U-Net and its visual features (encoded via CLIP) for cross-attention conditioning. LADI-VTON~\cite{morelli2023ladi}, instead, proposes to project the garment features into the CLIP text embedding space using textual inversion~\cite{gal2022textual}, thereby establishing a clear distinction between the pre-computed warping step and the core virtual try-on task. Despite these improvements, most existing methods remain confined to virtual try-on scenarios, limiting personalization to a single specific garment, which reduces flexibility in meeting diverse user needs.

Recent efforts have moved from standard virtual try-on to more generalized image synthesis tasks, such as multimodal fashion image editing~\cite{baldrati2023multimodal,baldrati2024multimodal,wang2024texfit} where the generative process is guided by multimodal information (\eg, text, garment sketches, and body poses) without explicit garment inputs. Nevertheless, these models struggle to generate high-quality garment details due to the absence of explicit visual conditioning. In this work, we aim to follow this research path, enabling users to condition the output on multiple input images and instructions, thereby providing greater flexibility and control. By retrieving multiple reference images, our approach enhances fidelity to input modalities and improves garment detail generation through a novel conditioning mechanism that effectively integrates multi-image visual inputs.

\begin{figure*}[t]
    \centering
    \includegraphics[width=0.96\linewidth]{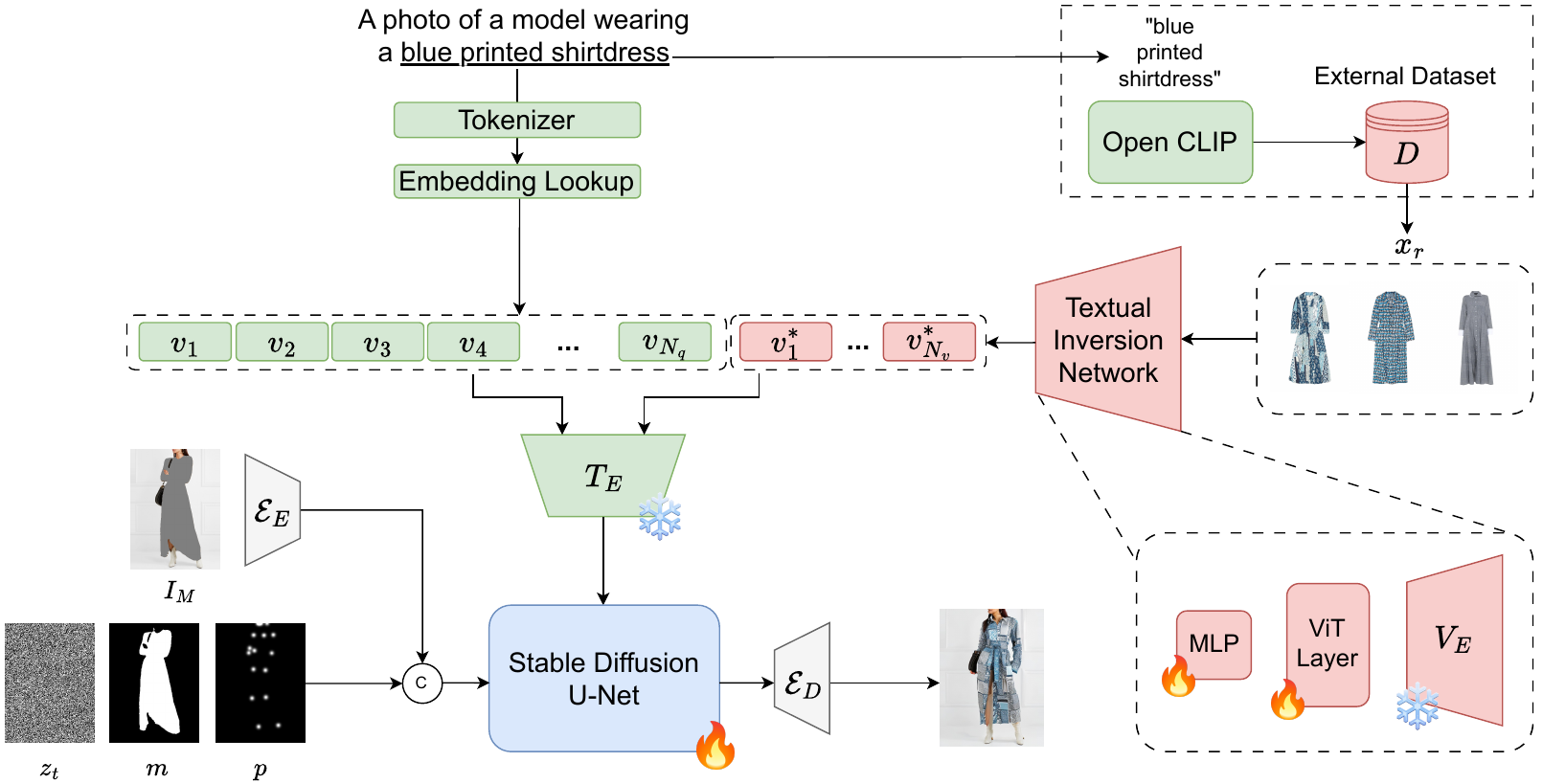}
    \vspace{-0.2cm}
    \caption{Overview of the proposed retrieval-augmented multimodal fashion image editing framework. The model leverages a diffusion-based inpainting pipeline, taking as input a masked reference image, a pose map, a binary mask indicating the editable region, and multimodal conditioning signals, including text descriptions and retrieved garments. Retrieved garments are projected into the CLIP textual space and combined with the textual embeddings to enhance the U-Net cross-attention mechanism. The U-Net iteratively denoises the latent representation over multiple steps, and the VAE decoder generates the final image.}
    \label{fig:model}
    \vspace{-0.3cm}
\end{figure*}

\tit{Retrieval-Augmented Generation}
In recent years, Retrieval-Augmented Generation (RAG) has emerged as a widely adopted paradigm in LLMs, particularly when training or fine-tuning models on task-specific or domain-specific data is infeasible. As outlined in~\cite{gao2023retrieval}, RAG-based approaches can be broadly categorized into three main types: (i) pre-training models using external datasets to enhance retrieval capabilities~\cite{rao2024raven}, (ii) fine-tuning pre-trained models directly on specific retrieval datasets~\cite{cheng2023uprise}, and (iii) zero-shot retrieval techniques that exploit external information without further training~\cite{shi2023replug}. This naive scheme can be extended into more advanced RAG frameworks, such as modular RAG-based pipelines~\cite{yu2022ragmodulargenerate}, which restructure retrieval and generation into flexible, independently adaptable components. Additionally, emerging model-driven retrieval strategies~\cite{asai2023selfrag,yu2024rankrag,ram2023context} focus on dynamically optimizing both the selection and utilization of retrieved information.

In addition to the LLM literature, retrieval-augmented architectures have extensively been explored in multimodal tasks such as image captioning~\cite{sarto2024towards,ramos2023smallcap,barraco2023little}, visual question answering~\cite{hu2023reveal,cocchi2025augmenting,caffagni2025recurrence}, and text-to-image generation~\cite{chen2022reimagen,blattmann2022retrieval,yasunaga2023retrieval}. However, there has been limited focus on applying these methods to specialized domains like fashion. In this work, we present, for the first time, a RAG-based solution directly integrated within the Stable Diffusion architecture, enabling it to leverage fine-grained information from multiple retrieved garments during the generation process.

\tit{Textual Inversion} 
Textual inversion~\cite{gal2022textual} is a powerful technique designed to encode image features into the CLIP textual embedding space. In its original formulation, images are tokenized and transformed via an embedding lookup mechanism to align with the dimensionality of the CLIP space. Building on this concept, Morelli~\etal~\cite{morelli2023ladi} re-adapts textual inversion to enable seamless integration of visual and textual conditioning by embedding pseudo-tokens into the CLIP space. This approach can be extended to support multiple concepts and styles, as demonstrated in~\cite{kumari2023multi}. Alternatively, StyleAligned~\cite{hertz2024style} introduces a novel method aimed at maintaining stylistic consistency across generated images by explicitly disentangling content from style. Along similar lines, the approach introduced in~\cite{alaluf2024app} proposes a technique for guiding image generation using an ``appearance'' image while preserving the geometric structure of a reference image.

To achieve greater control over geometric structure and style, plug-and-play methods such as ControlNet~\cite{zhang2023adding} and IP-Adapter~\cite{ye2023ip} have been explored. While these approaches are versatile, they also present some limitations: ControlNet often experiences a drop in image quality, likely due to the separate encoding of spatial inputs (\eg, pose) and latent noise vectors, while IP-Adapter lacks explicit geometric control, such as the ability to handle pose or masks. Some attempts to combine the strengths of both methods have been made, but these still suffer from the quality degradation issues inherent to ControlNet. Given these limitations, in this work, we adopt a modified textual inversion module to leverage retrieved information better, ensuring an improved balance between visual quality and controllability.
\section{Proposed Method}
\label{sec:method}

In this section, we introduce \ours, a retrieval-augmented generation framework designed for multimodal fashion image editing. Our approach builds on diffusion-based generative models, enhancing performance by incorporating knowledge from an external fashion database through garment retrieval. Unlike existing methods that rely solely on textual descriptions or pose information, \ours leverages both retrieved visual features and multimodal inputs to better guide the generative process. An overview of the proposed approach is depicted in Fig.~\ref{fig:model}.

\subsection{Multimodal Fashion Image Editing}
\tinytit{Task Definition} 
Recent research has explored the customization of fashion images through multimodal inputs, enabling selective editing of specific image regions, such as worn garments~\cite{baldrati2023multimodal,baldrati2024multimodal,wang2024texfit}. Unlike virtual try-on approaches, this setup does not require an input garment image to be virtually ``tried-on'', but conditions the generation through garment sketches, textual descriptions, or body pose maps. These inputs jointly guide the generative process to produce fashion images that align with the specified requirements. In this work, we enhance this paradigm by introducing additional conditioning through retrieved garments that align with the input textual description, offering finer control over the generation process and improving the quality of the synthesized outputs.

\tit{Architecture} 
Our approach leverages the Stable Diffusion inpainting pipeline as the core image generation backbone. The model takes as input a latent noise map $z$, a pose map $p$ representing the body keypoints, a binary mask $m$ indicating the region to be edited, the masked input image $I_M$ where the area to be inpainted is masked, the text captions and the corresponding $N_r$ retrieved garments. The diffusion U-Net is conditioned through the convolutional input and the cross-attention layers. We leverage the convolutional input for the latent noise map $z$ and the spatially aligned conditioning input (\ie, the pose map $p$, the binary mask $m$, and the masked input $I_M$), and the cross-attention modules for the others (\ie, the text captions and the relative $N_r$ retrieved garments). We encode the masked image $I_M$ into a low-dimensional latent representation using the VAE encoder $\mathcal{E}_E$. The spatial input is then formed by concatenating the latent noise, pose map, binary mask, and the encoded input image. During inference, the U-Net~\cite{ronneberger2015u} processes these inputs over $T$ denoising steps, and the VAE decoder $\mathcal{E}_D$ generates the final image. In parallel, the retrieved garments are projected into the same CLIP embedding space as the text captions and concatenated with the textual embeddings. The resulting embeddings are encoded by the CLIP text encoder and used to populate the key-value pairs in the U-Net cross-attention layers, enabling effective conditioning based on both garment retrieval and textual descriptions.

\subsection{Guiding the Generation via Retrieved Garments} 

\tinytit{CLIP-based Retrieval}
To enhance the generative process, we leverage $N_r$ retrieved garments extracted from an external database of fashion items. Specifically, given a textual description and an external database $\mathcal{D}$ of fashion images, we use OpenCLIP~\cite{Radford2021LearningTV,cherti2023reproducible} to retrieve relevant garments. First, the text is projected into the CLIP embedding space, and all images in $\mathcal{D}$ are encoded as CLIP image embeddings. Then, the cosine similarity between text and image embeddings is computed. The top-$N_r$ images with the highest similarity scores are selected as the retrieved garments and given as input to the generative model. Our framework supports varying the number of retrieved garments, and we evaluate performance for $N_r \in \{0, 1, 2, 3\}$.

\tit{Textual Inversion} 
Directly incorporating the visual information from the retrieved garments into the Stable Diffusion U-Net is not feasible, as the model expects textual embeddings as input. To bridge this modality gap, we adopt a modified textual inversion technique~\cite{gal2022textual}, which enables encoding images as pseudo-words compatible with the CLIP text encoder. Specifically, we follow the approach proposed in~\cite{morelli2023ladi}, where both visual and textual inputs are jointly utilized.

Given an image $x_r\in \mathbb{R}^{W \times H \times 3}$, the CLIP vision encoder $V_E$ extracts visual features that are subsequently projected into the CLIP textual space via a learnable network $F_\theta$. This network comprises a lightweight Vision Transformer (ViT)~\cite{dosovitskiy2021animage} followed by a Multi-Layer Perceptron (MLP), yielding $N_v$ visual embeddings. Formally, the visual embeddings are obtained as follows:
\begin{equation}
    \mathbf{v^*}=F_{\theta}(V_E(x_r)),
\end{equation}
where $\mathbf{v^*} \in \mathbb{R}^{N_v \times h_E}$ (\ie, $\mathbf{v^*}=\{v^*_1, v^*_2, v^*_3, \dots, v^*_{N_v}\}$) and $h_E$ denotes the CLIP embedding dimensionality. When $N_r > 1$, we extract $\mathbf{v^*}$ for each retrieved garment $x_r$ and concatenate all extracted embeddings.

In parallel, the input text is tokenized into $N_q$ tokens and encoded through an embedding-lookup layer, producing textual features $\mathbf{v} \in \mathbb{R}^{N_q \times h_E}$ (\ie, $\mathbf{v}=\{v_1, v_2, v_3, \dots, v_{N_q}\}$). The concatenated features $[\mathbf{v}; \mathbf{v^*}]$ are padded to form an input of length $N_L$, ensuring compatibility with the expected input dimension of the CLIP textual encoder. 

The resulting combined embeddings $x_c \in \mathbb{R}^{N_L \times h_E}$, eventually zero-padded, are fed into the CLIP text encoder of Stable Diffusion. The output from the CLIP encoder, $T_E(x_c)$, is subsequently utilized in the cross-attention modules of the model. Formally, the cross-attention mechanism is defined as: 
\begin{equation}
   \text{Attention}(\mathbf{Q}, \mathbf{K}, \mathbf{V}) = softmax\left(\frac{\mathbf{Q}\mathbf{K}^T}{\sqrt{d}}\right)\mathbf{V} 
\end{equation}
where $\mathbf{K} = \mathbf{W}_{K}  T_E(x_c)$ and $\mathbf{V} = \mathbf{W}_{V}  T_E(x_c)$ represent the key and value matrices derived from the combined embeddings (derived from the input text and the textual inverted visual features from retrieved garments), $\mathbf{Q}=\mathbf{W}_{Q} \gamma$ denotes the query matrix computed from the spatial inputs processed by the U-Net, and $d$ is a scaling factor. Here, $\mathbf{W}_Q$, $\mathbf{W}_K$, and $\mathbf{W}_V$ are learned matrices and $\gamma$ represents the spatial latent features at denoising step $t$. This approach enables fine-grained control over the generation process by integrating visual attributes from retrieved garments into the generative pipeline.

\begin{table*}[t]
\caption{Experimental comparison with state-of-the-art methods on paired and unpaired settings of the Dress Code multimodal test set~\cite{morelli2022dresscode,baldrati2023multimodal}. For each model, we specify whether it uses retrieval augmentation and report the number $N_r$ of retrieved items.}
\label{tab:dresscode}
\vspace{-0.1cm}
\footnotesize
\setlength{\tabcolsep}{.35em}
\resizebox{\linewidth}{!}{
\begin{tabular}{lccc cccccc c cccc}
\toprule
 & & & & \multicolumn{6}{c}{\textbf{Paired Setting}}  & & \multicolumn{4}{c}{\textbf{Unpaired Setting}} \\
\cmidrule{5-10} \cmidrule{12-15}
\textbf{Model} & \textbf{RAG} & $N_r$ & & \textbf{LPIPS} $\downarrow$ & \textbf{SSIM} $\uparrow$ & \textbf{FID} $\downarrow$ & \textbf{KID} $\downarrow$ & \textbf{CLIP-T} $\uparrow$ & \textbf{CLIP-I} $\uparrow$ & & \textbf{FID} $\downarrow$ & \textbf{KID} $\downarrow$ & \textbf{CLIP-T} $\uparrow$ & \textbf{CLIP-I} $\uparrow$ \\
\midrule
SD v2.1~\cite{rombach2022high} & \xmark & - & & 0.182 & 0.806 & 12.79 & 5.88 & 30.43 & - & & 14.14 & 6.58 & 29.44 & - \\
ControlNet~\cite{zhang2023adding} & \xmark & - & & 0.155 & 0.832 & 7.09 & 2.08 & 19.42 & - & & 8.25 & 2.58 & 19.21 & - \\
ControlNet~\cite{zhang2023adding}+IP~\cite{ye2023ip} & \cmark & 1 & & 0.158 & 0.833 & 9.17 & 3.49 & 27.54 & 53.70 & & 10.50 & 4.13 & 26.95 & 52.70 \\
\midrule 
MGD~\cite{baldrati2023multimodal} & \xmark & - & & - & - & 5.74 &  2.11 & \textbf{31.68} & - & & 7.73 & 2.82 & \textbf{30.04} & - \\
\rowcolor{ourcolor}
\textbf{\ours (Ours)} & \cmark & 1 & & 0.106 & 0.868 & 6.53 & 2.78 & 29.62 & \textbf{60.09} & & 7.12 & 2.25 & 26.74 & \textbf{56.84} \\
\rowcolor{ourcolor}
\textbf{\ours (Ours)} & \cmark & 2 & & 0.106 & 0.869 & 5.46 & 1.58 & 29.95 & 43.80 & & 6.91 & 2.17 & 27.77 & 41.66 \\
\rowcolor{ourcolor}
\textbf{\ours (Ours)} & \cmark & 3 & & \textbf{0.103} & \textbf{0.870} & \textbf{5.42} & \textbf{1.49} & 30.43 & 43.97 & & \textbf{6.88} & \textbf{2.11} & 27.97 & 41.77 \\
\bottomrule
\end{tabular}
}
\vspace{-0.3cm}
\end{table*} 

\subsection{Training} 
Given the input image $I \in \mathbb{R}^{H \times W \times 3}$ and the corresponding latent representation dimensions $h = H/8$ and $w = W/8$, we follow the standard Stable Diffusion training framework. Specifically, we corrupt the image latent representation by adding Gaussian noise $\epsilon \sim \mathcal{N}(0,1)$ according to a given timestep and train the U-Net to estimate the noise, conditioned on both global and spatial inputs. 

For spatial conditioning, we resize the binary mask $M \in \mathbb{R}^{H \times W \times 1}$ and the pose map $P \in \mathbb{R}^{H \times W \times 18}$ respectively to $m \in \mathbb{R}^{h \times w \times 1}$ and $p \in \mathbb{R}^{h \times w \times 18}$, where the 18 channels represent the human body keypoints heatmaps. Also, we encode the masked input image $I_M = I \cdot M$ via the VAE encoder $\mathcal{E}_E$ and then concatenate along the channel dimension the noise latent variable $z_t \in \mathbb{R}^{h \times w \times 4}$ with the previously computed inputs forming the combined spatial input:

\begin{equation}
    \gamma = [z_t; m; \mathcal{E}(I_M); p] \in \mathbb{R}^{h\times w \times C}
\end{equation}
where $C$ is the total number of channels after concatenation (\ie, $27$ in our setting). 

The global conditioning is provided by the embeddings $\psi = T_E(x_c) \in \mathbb{R}^{N_L \times h_E}$, where $T_E$ denotes the CLIP text encoder, and $x_c$ represents the concatenated visual and textual embeddings as described in the previous section. The overall training objective minimizes the denoising error using the following loss function:
\begin{equation}
     \mathcal{L} = \mathbb{E}_{\epsilon \sim \mathcal{N}(0,1), t, \mathcal{E}(I_M), x_c, m, p} ||\epsilon - \epsilon_{\theta}(\gamma, \psi) ||_2^2
\end{equation}
where $\epsilon_\theta(\gamma, \psi)$ denotes the U-Net noise prediction at time step $t$, conditioned on the spatial input $\gamma$ and global conditioning $\psi$. This formulation ensures that the model effectively learns to denoise the corrupted latent representation while leveraging both spatial and global information.
\section{Experiments}
\label{sec:experiments}

\subsection{Implementation Details} 
\tinytit{Architecture} 
We build our method upon the Stable Diffusion v2.1~\cite{rombach2022high} inpainting model\footnote{\url{https://huggingface.co/stabilityai/stable-diffusion-2-inpainting}}. We employ OpenCLIP ViT-H/14~\cite{Radford2021LearningTV,cherti2023reproducible} as visual and textual encoder, in which the token length is $N_L=77$ and the token features channels are $h_E=1,024$. The spatial input dimensions are $H=512$ and $W=384$, scaled by $8$ per spatial dimension when projected in latent space. A fixed prompt (\ie, ``\texttt{A photo of a model wearing a}'') is prepended to the input text, followed by the textual inverted visual features of the retrieved garments (with $N_v=16$ for each retrieved item). As a retrieval model, we employ the ViT-L/14 version of OpenCLIP to compute the correlation between textual and visual features.

\tit{Training Details}
Our model is trained with a two-stage training strategy. In the first stage, we only train the network $F_{\theta}$ for 200k steps to learn how to map the visual features of retrieved garments to the CLIP textual space. In the second stage, following~\cite{baldrati2023multimodal}, we add zero-initialized modules in the first U-Net convolution to take as an additional input the pose map $p$ and fine-tune both the U-Net and $F_{\theta}$ for 120k steps. We use a variable number of retrieved items $N_r$ sampled uniformly between 0 and 3 with equal probability during training. Also, we employ the multi-condition classifier-free guidance proposed in~\cite{avrahami2022spatext} to speed up the inference. All experiments are done on a single NVIDIA A100 GPU, with AdamW~\cite{loshchilov2019decoupled} as the optimizer and learning rate $1 \times 10^{-5}$.

\begin{figure*}[t]
\begin{minipage}[t]{0.153\linewidth}
\centering
\footnotesize{\textbf{Reference}}\\
\footnotesize{\textbf{Model}}

\vspace{0.161cm}
\includegraphics[width=1.\linewidth]{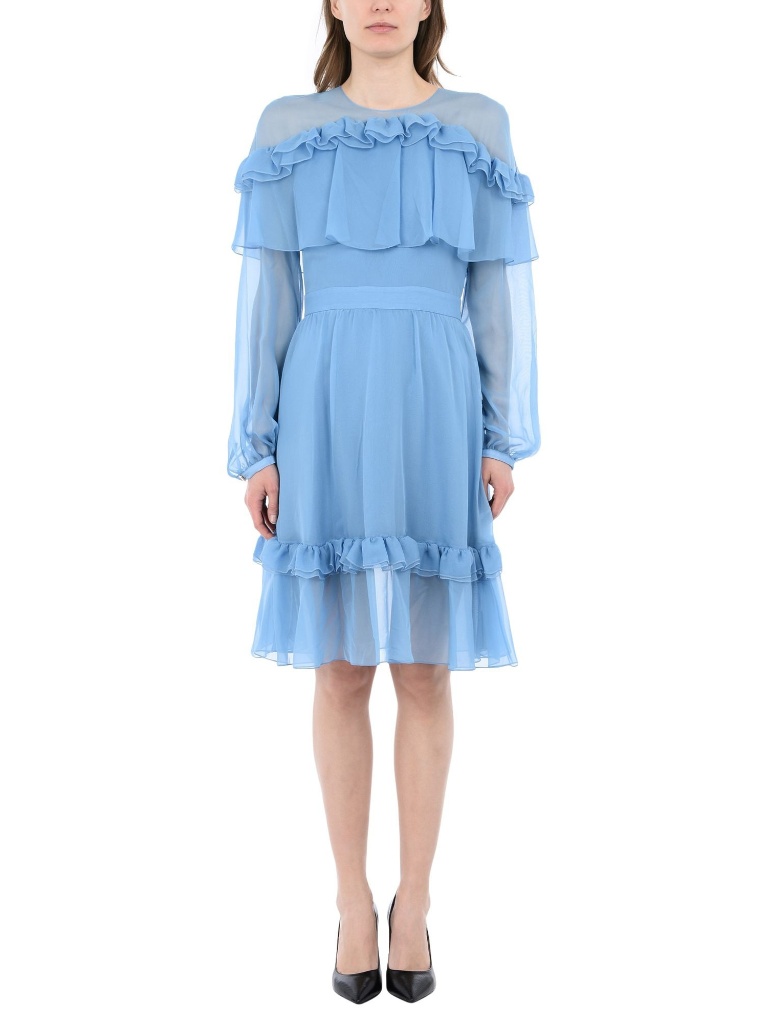}
\end{minipage}
\hspace{0.01cm}
\begin{minipage}[t]{0.17\linewidth}
\centering
\footnotesize{\textbf{Input}}\\
\footnotesize{\textbf{Text}}

\vspace{0.5cm}
\footnotesize{long white dress, rouched front, solid white and floor length}
\end{minipage}
\hspace{0.01cm}
\begin{minipage}[t]{0.152\linewidth}
\centering
\footnotesize{\textbf{SD v2.1}}\\
\footnotesize{\textbf{\cite{rombach2022high}}}

\vspace{0.1cm}
\includegraphics[width=1.\linewidth]{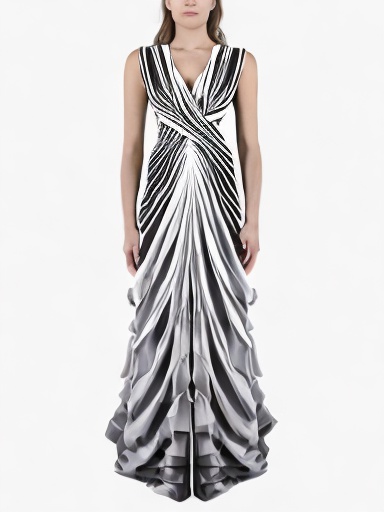}
\end{minipage}
\hspace{0.01cm}
\begin{minipage}[t]{0.152\linewidth}
\centering
\footnotesize{\textbf{ControlNet}}\\
\footnotesize{\textbf{+IP~\cite{zhang2023adding,ye2023ip}}}

\vspace{0.1cm}
\includegraphics[width=1.\linewidth]{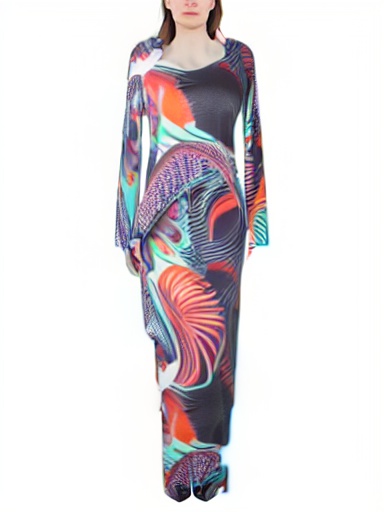}
\end{minipage}
\hspace{0.01cm}
\begin{minipage}[t]{0.152\linewidth}
\centering
\footnotesize{\textbf{MGD}}\\
\footnotesize{\textbf{\cite{baldrati2023multimodal}}}

\vspace{0.1cm}
\includegraphics[width=1.\linewidth]{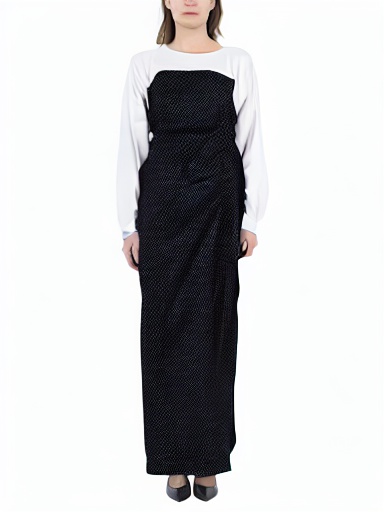}
\end{minipage}
\hspace{0.01cm}
\begin{minipage}[t]{0.152\linewidth}
\centering
\footnotesize{\textbf{Fashion-}}\\
\footnotesize{\textbf{RAG (Ours)}}

\vspace{0.1cm}
\includegraphics[width=1.\linewidth]{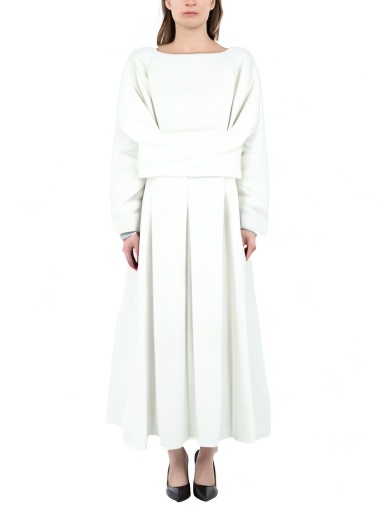}
\end{minipage}

\vspace{0.05cm}
\begin{minipage}[t]{0.152\linewidth}
\vspace{0pt}\centering
\includegraphics[width=1.\linewidth]{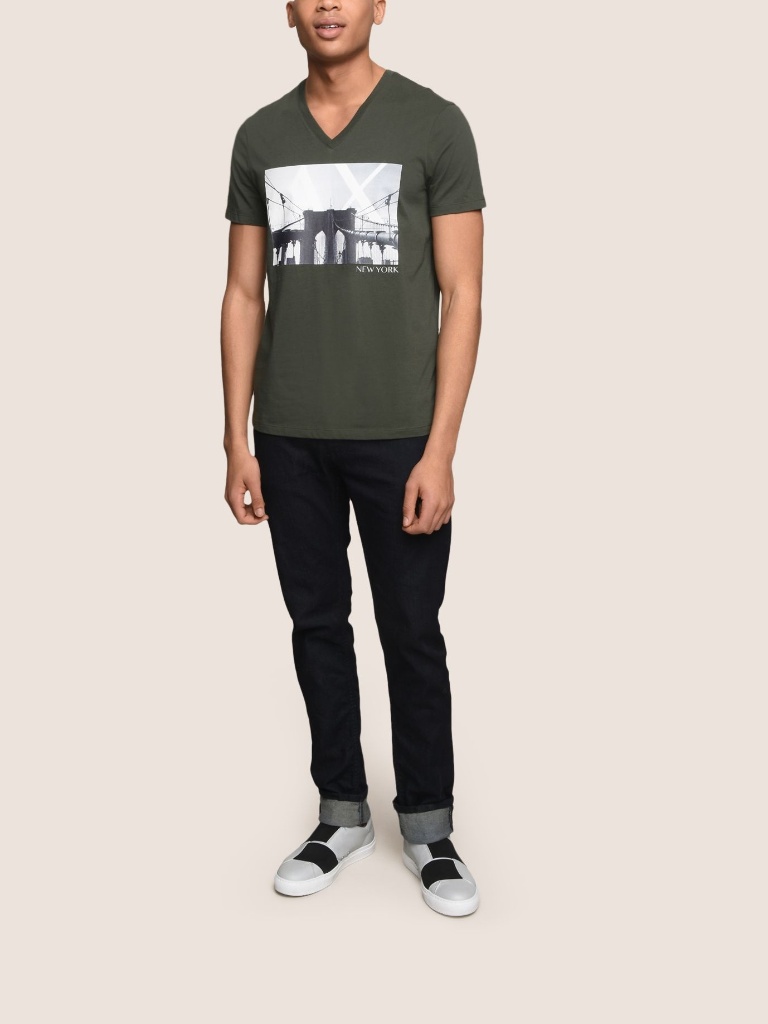}
\end{minipage}
\hspace{0.01cm}
\begin{minipage}[t]{0.17\linewidth}
\vspace{0.5cm}\centering
\footnotesize{3-stripe short sleeve polo shirt, golf-style shirt, white short sleeved polo
}
\end{minipage}
\hspace{0.01cm}
\begin{minipage}[t]{0.152\linewidth}
\vspace{0pt}\centering
\includegraphics[width=1.\linewidth]{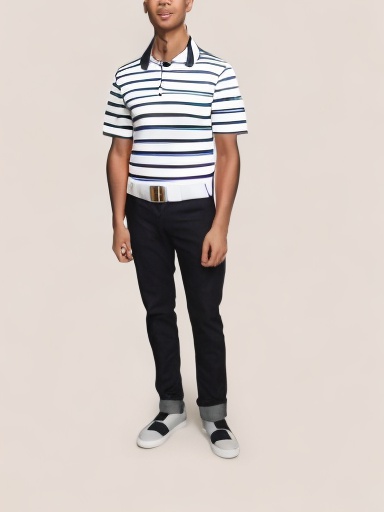}
\end{minipage}
\hspace{0.01cm}
\begin{minipage}[t]{0.152\linewidth}
\vspace{0pt}\centering
\includegraphics[width=1.\linewidth]{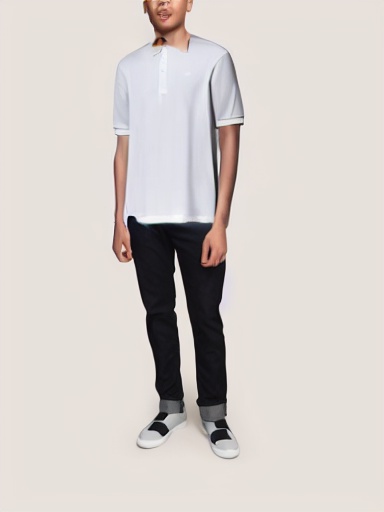}
\end{minipage}
\hspace{0.01cm}
\begin{minipage}[t]{0.152\linewidth}
\vspace{0pt}\centering
\includegraphics[width=1.\linewidth]{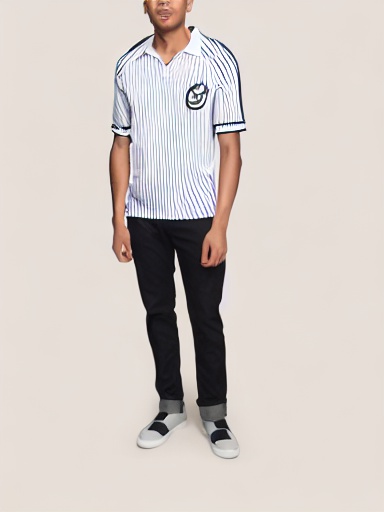}
\end{minipage}
\hspace{0.01cm}
\begin{minipage}[t]{0.152\linewidth}
\vspace{0pt}\centering
\includegraphics[width=1.\linewidth]{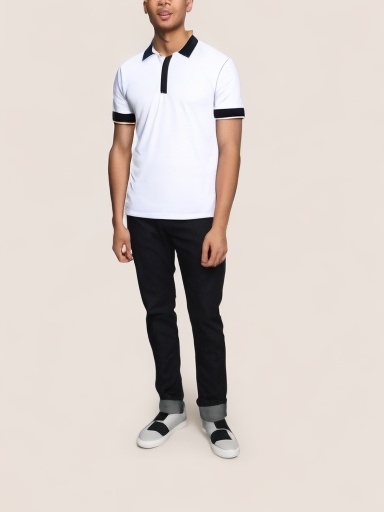}
\end{minipage}

\vspace{0.05cm}
\begin{minipage}[t]{0.152\linewidth}
\vspace{0pt}\centering
\includegraphics[width=1.\linewidth]{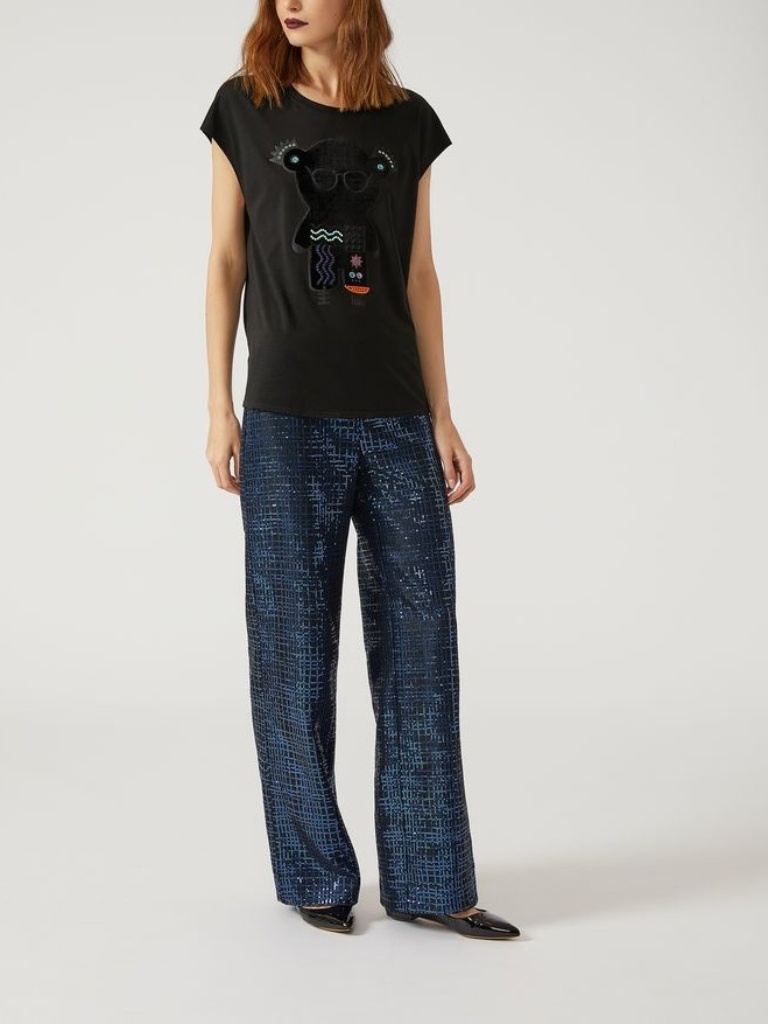}
\end{minipage}
\hspace{0.01cm}
\begin{minipage}[t]{0.171794\linewidth}
\vspace{0.5cm}\centering
\footnotesize{belted top, burnt orange top, orange blouse}
\end{minipage}
\hspace{0.01cm}
\begin{minipage}[t]{0.152\linewidth}
\vspace{0pt}\centering
\includegraphics[width=1.\linewidth]{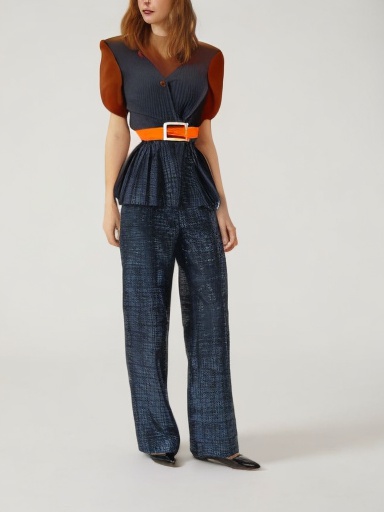}
\end{minipage}
\hspace{0.01cm}
\begin{minipage}[t]{0.152\linewidth}
\vspace{0pt}\centering
\includegraphics[width=1.\linewidth]{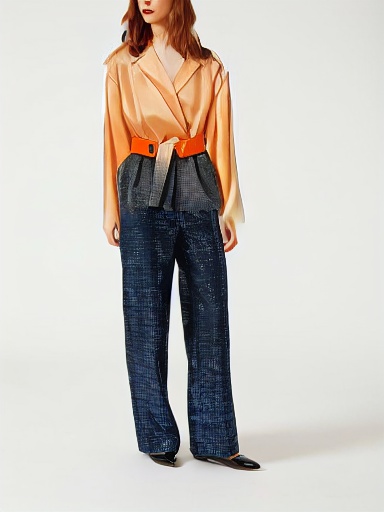}
\end{minipage}
\hspace{0.01cm}
\begin{minipage}[t]{0.152\linewidth}
\vspace{0pt}\centering
\includegraphics[width=1.\linewidth]{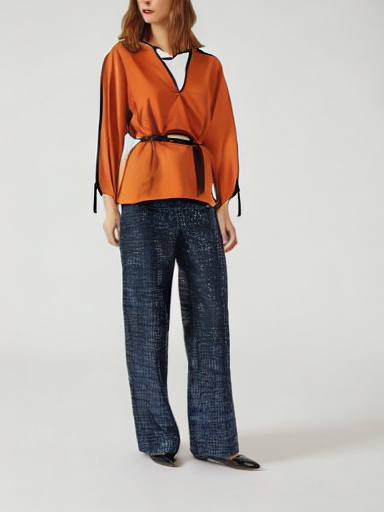}
\end{minipage}
\hspace{0.01cm}
\begin{minipage}[t]{0.152\linewidth}
\vspace{0pt}\centering
\includegraphics[width=1.\linewidth]{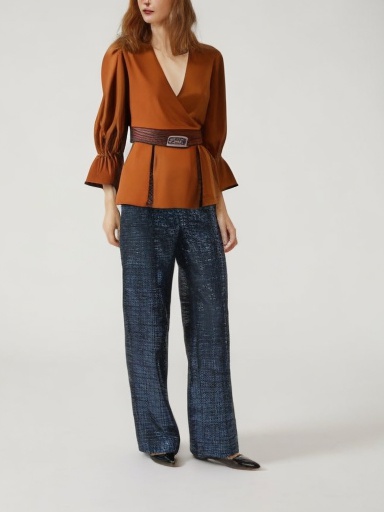}
\end{minipage}
\vspace{-0.2cm}
\caption{Visual comparison between our work (latest to the right) with other multimodal competitors. Previous methods struggle to adhere to some types textual inputs, and we show how this method sticks to them, such as rendering correct garment length (raw 1) without artifacts, reproducing fine-grained pattern and fabric textures (raw 2), generate correct color and additional objects (such as the belt, raw 3).}
\label{fig:comparison}
\vspace{-0.3cm}
\end{figure*}

\subsection{Dataset and Evaluation Metrics} 
We conduct our experiments on the Dress Code multimodal dataset~\cite{morelli2022dresscode,baldrati2023multimodal}, consisting of 53k image pairs. Each pair features an image of a garment and the corresponding reference model wearing it, categorized into three main categories: upper-body, lower-body, and full-body clothing. The dataset is split into training and test sets, comprising 48k and 5.4k samples, respectively. Each sample is supplemented with multimodal annotations, including a textual description and a pose map representing human body keypoints. The textual descriptions are generated by concatenating three noun phrases, providing concise and detailed captions of the garments. We exclusively utilize reference model images and their corresponding multimodal annotations for model training. In contrast, garment images from the training set are employed to build the external database of fashion items, from which retrieved garments are selected.

To quantitatively evaluate our model, we leverage a set of evaluation metrics to measure the coherence and realism of the generated images. Specifically, we measure the Learned Perceptual Image Patch Similarity (LPIPS)~\cite{zhang2018unreasonable} and the Structural Similarity (SSIM)~\cite{wang2004image} on the dataset paired setting, to evaluate how closely the generated images align with the ground-truth ones. To measure realism, we compute Fréchet Inception Distance (FID)~\cite{heusel2017gans} and Kernel Inception Distance (KID)~\cite{binkowski2018demystifying} in both paired and unpaired settings\footnote{The paired setting of the dataset consists of reference model images accompanied by multimodal annotations corresponding to the garment originally worn by the model. Conversely, the unpaired setting comprises reference model images paired with multimodal annotations from a different garment.}. To assess the coherence of the generated garment with respect to the textual constraint, we measure the semantic alignment between the input text and the generated images, using cosine similarity scores between the CLIP embeddings of the generated image and the input text (denoted as CLIP-T). Since CLIP-T does not consider the visual conditioning from retrieved images, we also compute the similarity between the CLIP embeddings of each generated image and each relative retrieved item (denoted as CLIP-I). The final CLIP-I score is obtained by averaging the individual CLIP-I scores of all $N_r$ retrieved items.

\subsection{Experimental Results} 
\tinytit{Baseline and Competitors}
We evaluate our approach against state-of-the-art methods for multimodal fashion image editing. Specifically, we compare it with the vanilla Stable Diffusion v2.1 inpainting model~\cite{rombach2022high} and ControlNet~\cite{zhang2023adding}, both conditioned solely on textual descriptions. Additionally, we include MGD~\cite{baldrati2023multimodal}, which originally introduced the task and employs conditioning on text, body pose, and garment sketches. Finally, we design a retrieval-augmented competitor by adapting ControlNet with IP-Adapter~\cite{ye2023ip}, enabling it to take as input a single retrieved garment along with textual descriptions.

\begin{table*}[t]
\centering
\caption{Ablation study results on the effect of the number of noun phrases ($N_c$) and the number of retrieved garments ($N_r$) on model performance. Results are reported on paired and unpaired settings of the Dress Code multimodal test set~\cite{morelli2022dresscode,baldrati2023multimodal}.}
\label{tab:ablations}
\vspace{-0.1cm}
\footnotesize
\setlength{\tabcolsep}{.38em}
\resizebox{0.84\linewidth}{!}{
\begin{tabular}{cccc cccccc c cccc}
\toprule
 & & & & \multicolumn{6}{c}{\textbf{Paired Setting}}  & & \multicolumn{4}{c}{\textbf{Unpaired Setting}} \\
\cmidrule{5-10} \cmidrule{12-15}
\textbf{RAG} & $\mathbf{N_c}$ & $\mathbf{N_r}$ & & \textbf{LPIPS} $\downarrow$ & \textbf{SSIM} $\uparrow$ & \textbf{FID} $\downarrow$ & \textbf{KID} $\downarrow$ & \textbf{CLIP-T} $\uparrow$ & \textbf{CLIP-I} $\uparrow$ & & \textbf{FID} $\downarrow$ & \textbf{KID} $\downarrow$ & \textbf{CLIP-T} $\uparrow$ & \textbf{CLIP-I} $\uparrow$ \\
\midrule
\xmark & 1 & 0 & & 0.117 & 0.862 & 6.87 & 2.23 & \textbf{30.15} & - & & 8.62 & 2.98 & \textbf{27.83} & -  \\
\cmark & 1 & 1 & & 0.117 & 0.864 & 6.09 & 1.83 & 28.53 & \textbf{59.39} & & 7.18 & 2.33 & 26.37 & \textbf{56.53} \\
\cmark & 1 & 2 & & 0.115 & 0.865 & \textbf{5.94} & \textbf{1.75} & 29.07 & 42.96 & & 7.12 & \textbf{2.22} & 27.20 & 41.11 \\
\rowcolor{ourcolor}
\cmark & 1 & 3 & & \textbf{0.114} & \textbf{0.866} & 5.99 & 1.87 & 29.27 & 43.43 & & \textbf{7.11} & 2.27 & 27.54 & 41.63 \\
\midrule
\xmark & 2 & 0 & & 0.109 & 0.866 & 6.07 & 1.92 & \textbf{31.54} & - & & 7.84 & 2.55 & \textbf{29.12} & - \\
\cmark & 2 & 1 & & 0.109 & 0.867 & 5.63 & 1.58 & 29.17 & \underline{\textbf{60.37}} & & 7.06 & 2.20 & 27.06 & \textbf{56.63} \\
\cmark & 2 & 2 & & 0.107 & 0.868 & 5.50 & \textbf{1.52} & 29.69 & 43.29 & & 6.94 & \textbf{2.12} & 27.45 & 41.08 \\
\rowcolor{ourcolor}
\cmark & 2 & 3 & & \textbf{0.106} & \textbf{0.869} & \textbf{5.46} & 1.58 & 29.95 & 43.80 & &\textbf{6.91} & 2.17 & 27.77 & 41.66 \\
\midrule
\xmark & 3 & 0 & & 0.106 & 0.869 & 5.90 & 1.81 & \underline{\textbf{32.19}} & - & & 7.42 & 2.40 & \underline{\textbf{29.77}} & - \\
\cmark & 3 & 1 & & 0.106 & 0.868 & 5.56 & 1.67 & 29.62 & \textbf{60.09} & & 7.12 & 2.25 & 26.74 & \underline{\textbf{56.84}} \\
\cmark & 3 & 2 & & 0.104 & 0.869 & 5.48 & 1.55 & 30.21 & 43.54 & & 6.99 & 2.15 & 27.60 & 41.11 \\
\rowcolor{ourcolor}
\cmark & 3 & 3 & & \underline{\textbf{0.103}} & \underline{\textbf{0.870}} & \underline{\textbf{5.42}} & \underline{\textbf{1.49}} & 30.43 & 43.97 & & \underline{\textbf{6.88}} & \underline{\textbf{2.11}} & 27.97 & 41.77 \\
\bottomrule
\end{tabular}
}
\vspace{-0.3cm}
\end{table*} 

\tit{Results on the Multimodal Dress Code Dataset}
Table~\ref{tab:dresscode} presents quantitative results on both paired and unpaired settings of the Dress Code multimodal test set~\cite{morelli2022dresscode,baldrati2023multimodal}. As it can be seen, the proposed \ours approach consistently outperforms state-of-the-art methods, including ControlNet~\cite{zhang2023adding} and MGD~\cite{baldrati2023multimodal}, across various metrics. In particular, \ours achieves superior performance in terms of FID and KID, indicating enhanced image fidelity and realism. Despite incorporating retrieval-augmented generation with up to three retrieved items, our method maintains competitive CLIP-T scores compared to MGD, underscoring its robust semantic alignment capabilities. Moreover, \ours demonstrates a notable improvement in CLIP-I, significantly surpassing ControlNet+IP~\cite{zhang2023adding,ye2023ip} and highlighting its effectiveness in integrating visual information from retrieved garments.

As expected, conditioning the generation on more than one retrieved item generally results in lower CLIP-I scores. This is due to the difficulty of blending visual features from multiple garments, which can reduce similarity to any single garment. Nonetheless, all other metrics benefit significantly from using multiple retrieved elements, demonstrating the effectiveness of retrieval-augmented generation in enhancing both the coherence and overall quality of the generated outputs.

Qualitative results are presented in Fig.~\ref{fig:comparison}, where we compare images generated by our model with those produced by Stable Diffusion, ControlNet+IP, and MGD. Notably, \ours generates more realistic images that better capture and reflect the visual details described in the input text, further confirming the impact of enhancing the generative process with retrieved elements.

\tit{Ablation Studies} 
In Table~\ref{tab:ablations}, we investigate the impact of the number of noun phrases ($N_c$) and retrieved images ($N_r$) on model performance. The optimal configuration for $N_c$ is 3, as it yields more detailed and accurate garment descriptions compared to $N_c=1$ or $N_c=2$. Regarding the number of retrieved garments, performance improves across FID, KID, and CLIP-T as $N_r$ increases, highlighting the advantages of incorporating multiple retrieved items. As previously observed, the highest CLIP-I score is achieved when $N_r=1$, as the generation process focuses solely on a single retrieved item. Notably, when retrieval augmentation is disabled (\ie, $N_r=0$), performance significantly degrades, particularly in terms of generation realism, further validating the effectiveness of our retrieval-augmented generation framework.

\begin{figure}[t]
\begin{minipage}[t]{0.18\linewidth}
\centering
\footnotesize{\textbf{Retrieved}}\\
\footnotesize{\textbf{(top-1)}}

\vspace{0.1cm}
\includegraphics[width=1.\linewidth]{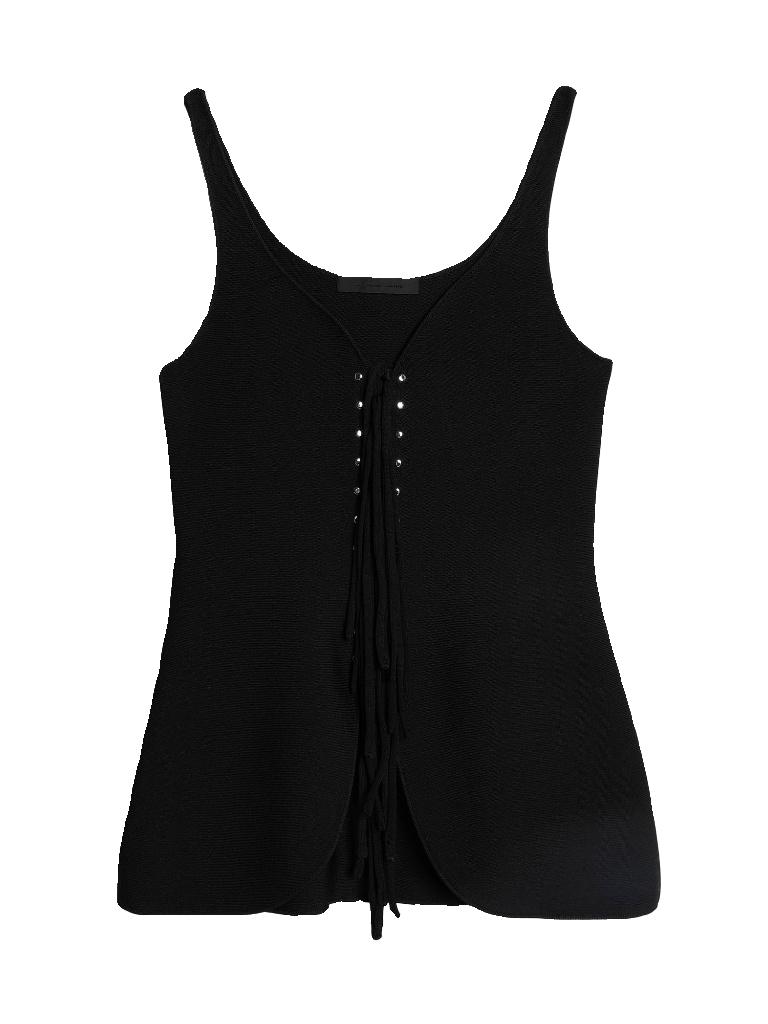}
\end{minipage}
\hspace{0.01cm}
\begin{minipage}[t]{0.18\linewidth}
\centering
\footnotesize{\textbf{Retrieved}}\\
\footnotesize{\textbf{(top-2)}}

\vspace{0.1cm}
\includegraphics[width=1.\linewidth]{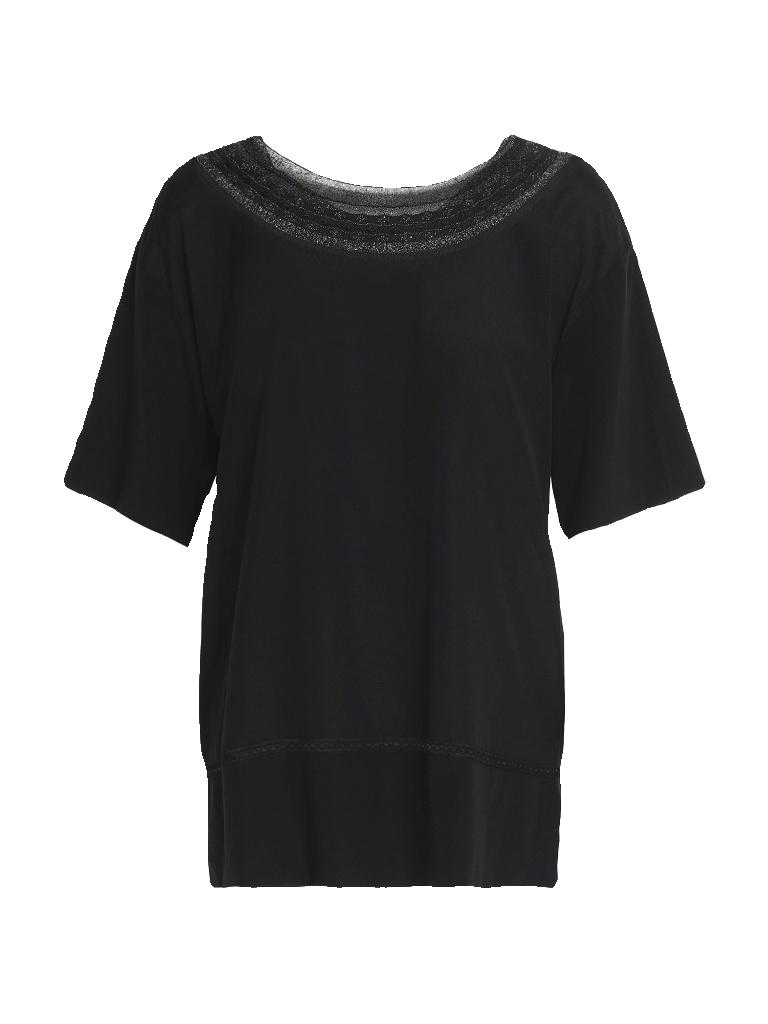}
\end{minipage}
\hspace{0.01cm}
\begin{minipage}[t]{0.18\linewidth}
\centering
\footnotesize{\textbf{Retrieved}}\\
\footnotesize{\textbf{(top-3)}}

\vspace{0.1cm}
\includegraphics[width=1.\linewidth]{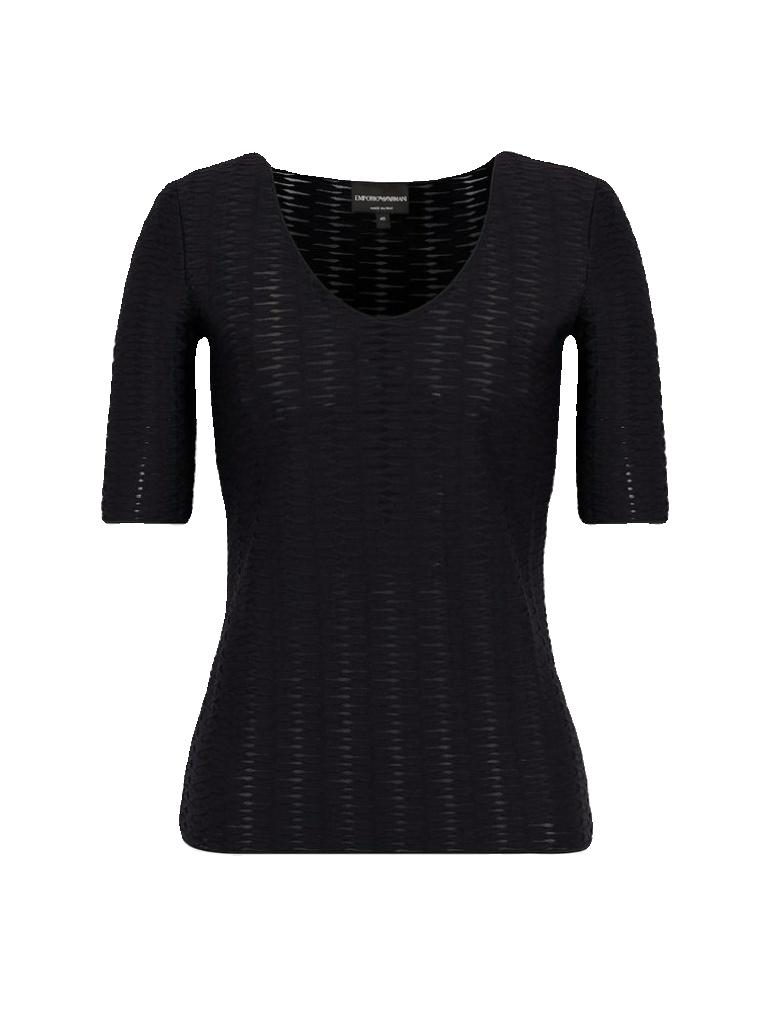}
\end{minipage}
\hspace{0.01cm}
\begin{minipage}[t]{0.18\linewidth}
\centering
\footnotesize{\textbf{w/o RAG}}\\
\footnotesize{($N_r=0$)}

\vspace{0.1cm}
\includegraphics[width=1.\linewidth]{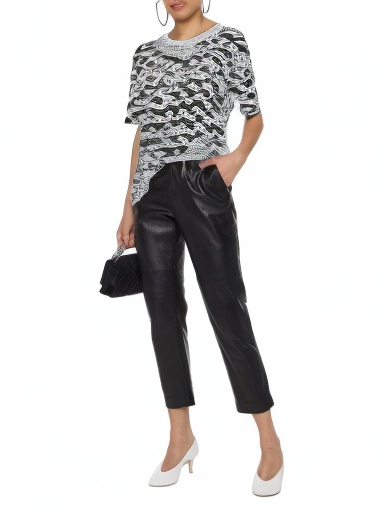}
\end{minipage}
\hspace{0.01cm}
\begin{minipage}[t]{0.18\linewidth}
\centering
\footnotesize{\textbf{w/ RAG}}\\
\footnotesize{($N_r=3$)}

\vspace{0.1cm}
\includegraphics[width=1.\linewidth]{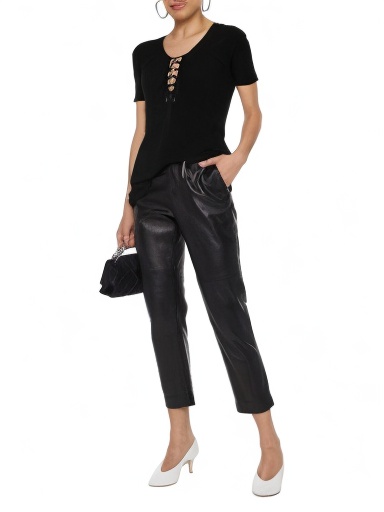}
\end{minipage}

\vspace{0.05cm}
\begin{minipage}[t]{0.18\linewidth}
\includegraphics[width=1.\linewidth]{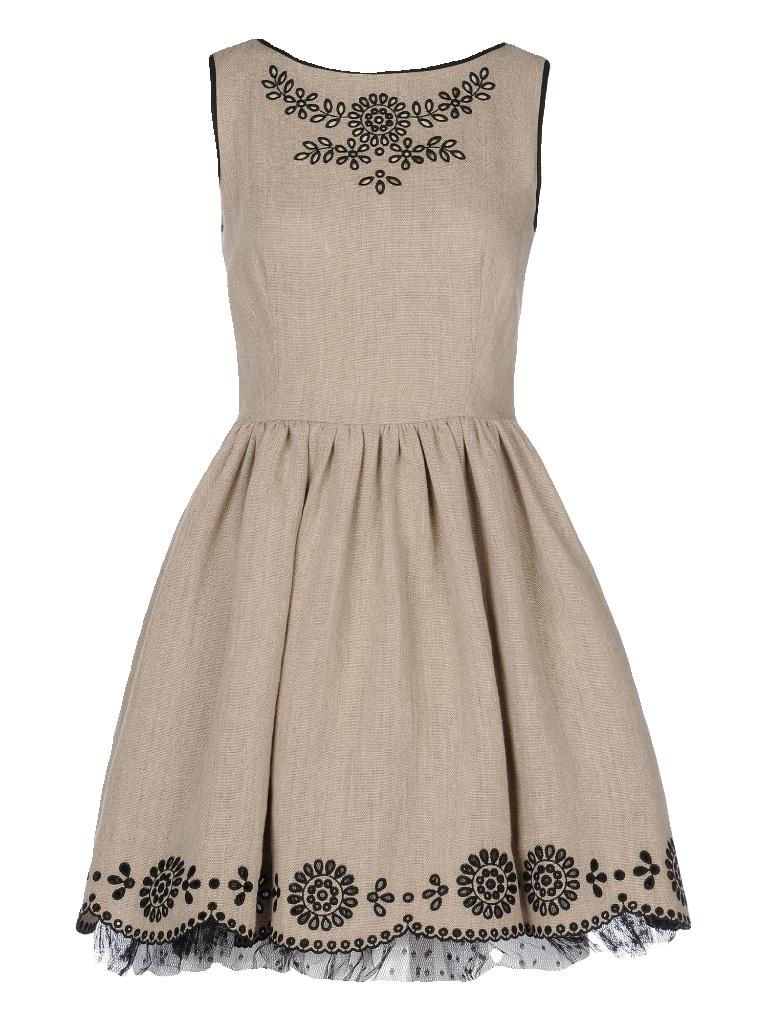}
\end{minipage}
\hspace{0.01cm}
\begin{minipage}[t]{0.18\linewidth}
\includegraphics[width=1.\linewidth]{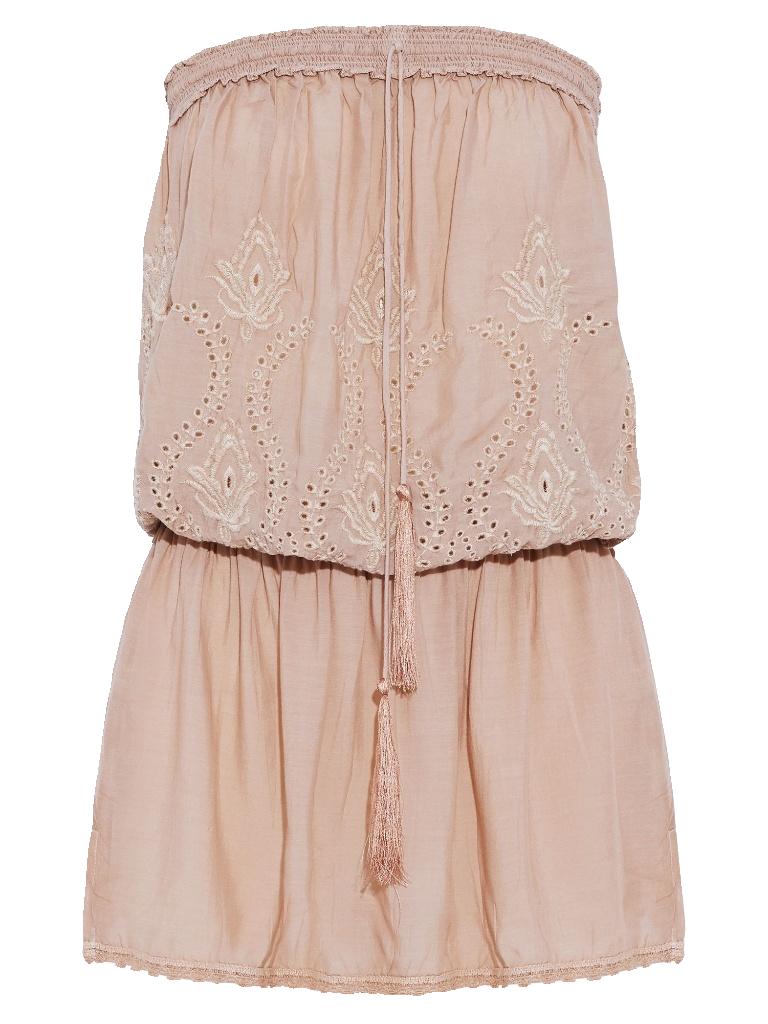}
\end{minipage}
\hspace{0.01cm}
\begin{minipage}[t]{0.18\linewidth}
\includegraphics[width=1.\linewidth]{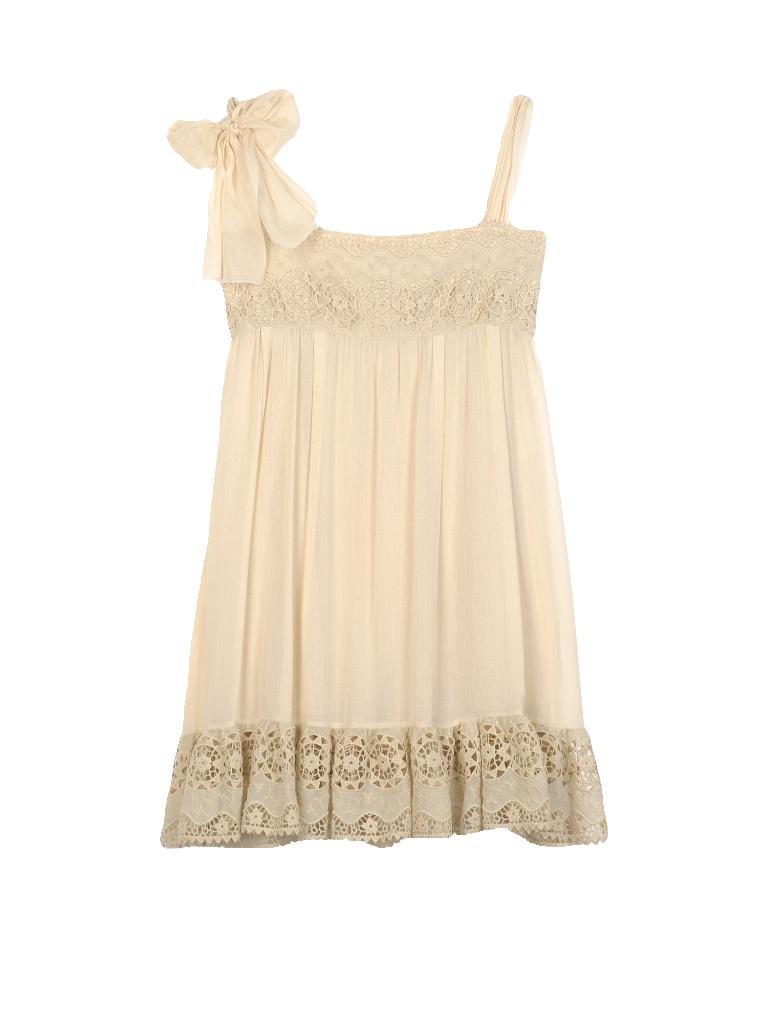}
\end{minipage}
\hspace{0.01cm}
\begin{minipage}[t]{0.18\linewidth}
\includegraphics[width=1.\linewidth]{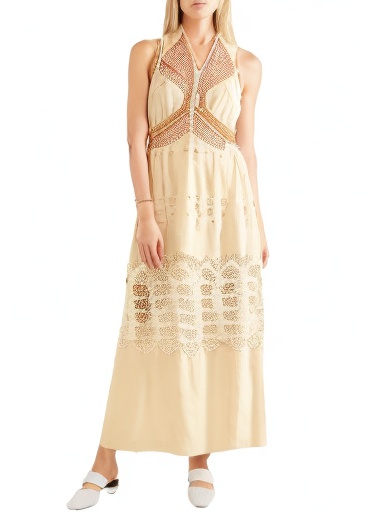}
\end{minipage}
\hspace{0.01cm}
\begin{minipage}[t]{0.18\linewidth}
\includegraphics[width=1.\linewidth]{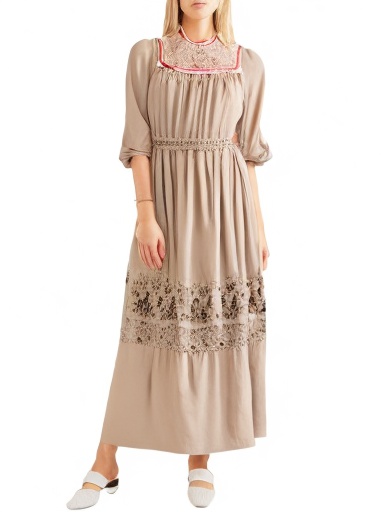}
\end{minipage}

\vspace{0.05cm}
\begin{minipage}[t]{0.18\linewidth}
\includegraphics[width=1.\linewidth]{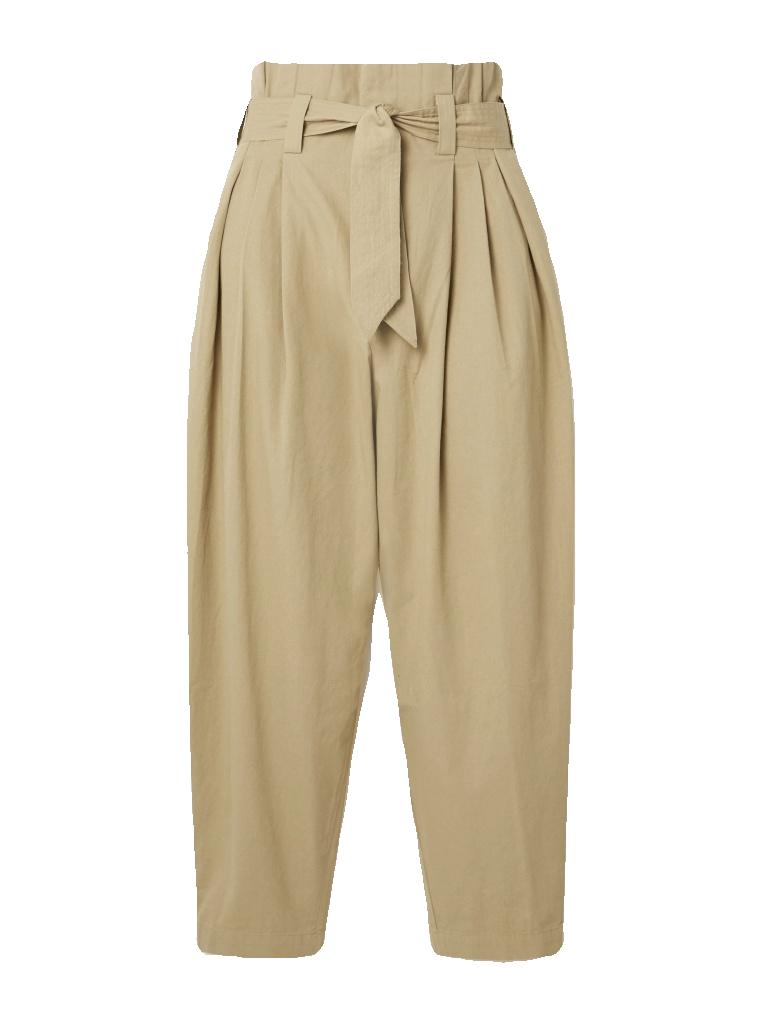}
\end{minipage}
\hspace{0.01cm}
\begin{minipage}[t]{0.18\linewidth}
\includegraphics[width=1.\linewidth]{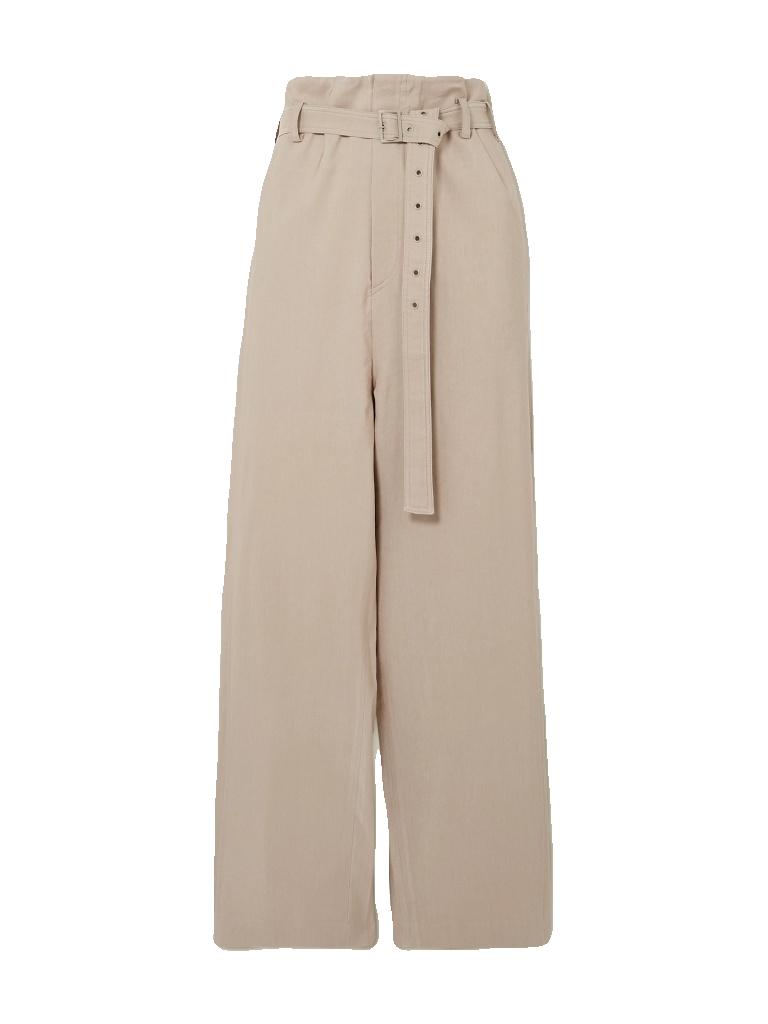}
\end{minipage}
\hspace{0.01cm}
\begin{minipage}[t]{0.18\linewidth}
\includegraphics[width=1.\linewidth]{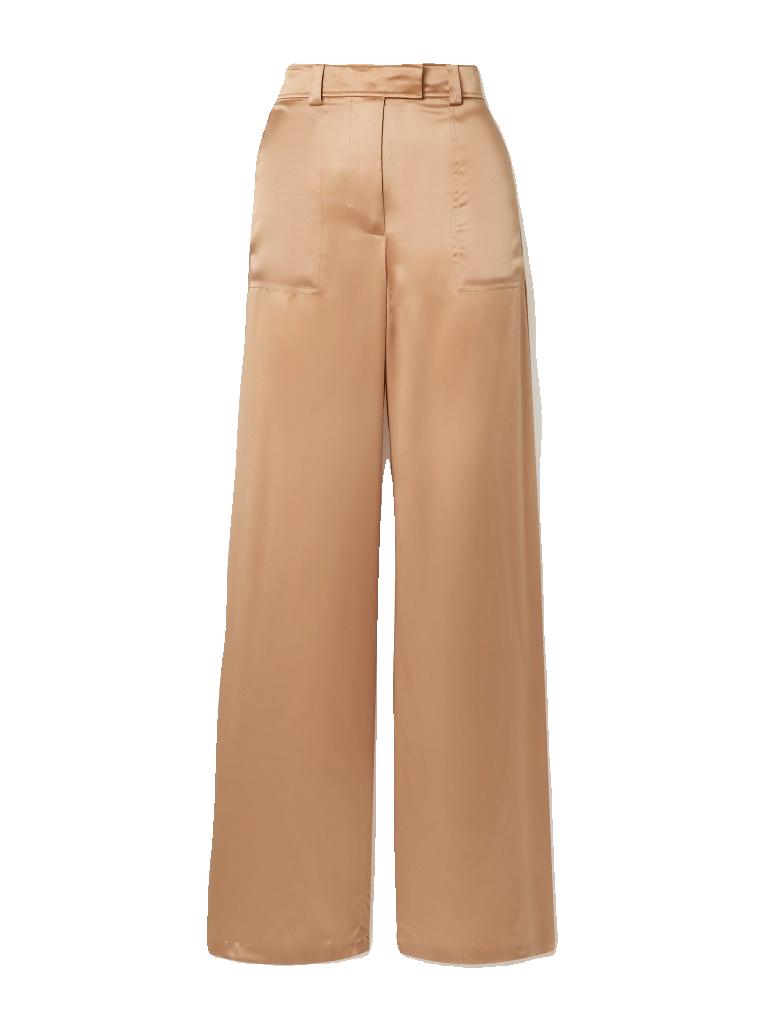}
\end{minipage}
\hspace{0.01cm}
\begin{minipage}[t]{0.18\linewidth}
\includegraphics[width=1.\linewidth]{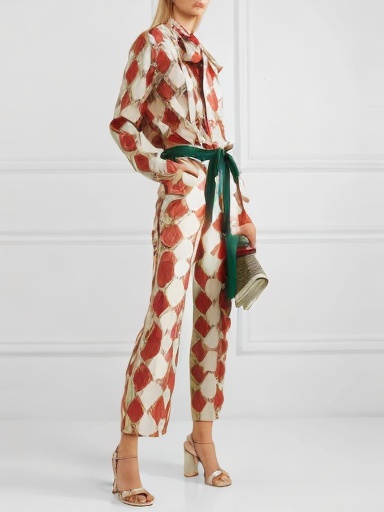}
\end{minipage}
\hspace{0.01cm}
\begin{minipage}[t]{0.18\linewidth}
\includegraphics[width=1.\linewidth]{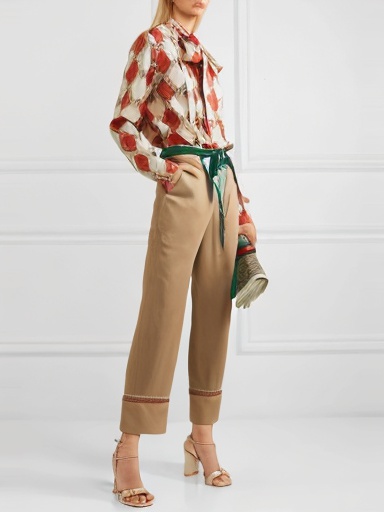}
\end{minipage}

\vspace{-0.25cm}
\caption{Qualitative comparison between images generated with and without retrieval augmentation, along with the top-3 retrieved garments.}
\label{fig:rag}
\vspace{-0.3cm}
\end{figure}

A qualitative comparison with and without retrieval augmentation is provided in Fig.~\ref{fig:rag}. As it can be seen, our approach can effectively combine visual features from multiple retrieved garments, resulting in highly realistic images that adhere closely to the provided conditioning.

\begin{figure}[t]
\begin{minipage}[t]{0.18\linewidth}
\centering
\footnotesize{\textbf{Reference}}\\
\footnotesize{\textbf{Model}}

\vspace{0.14cm}
\includegraphics[width=1.\linewidth]{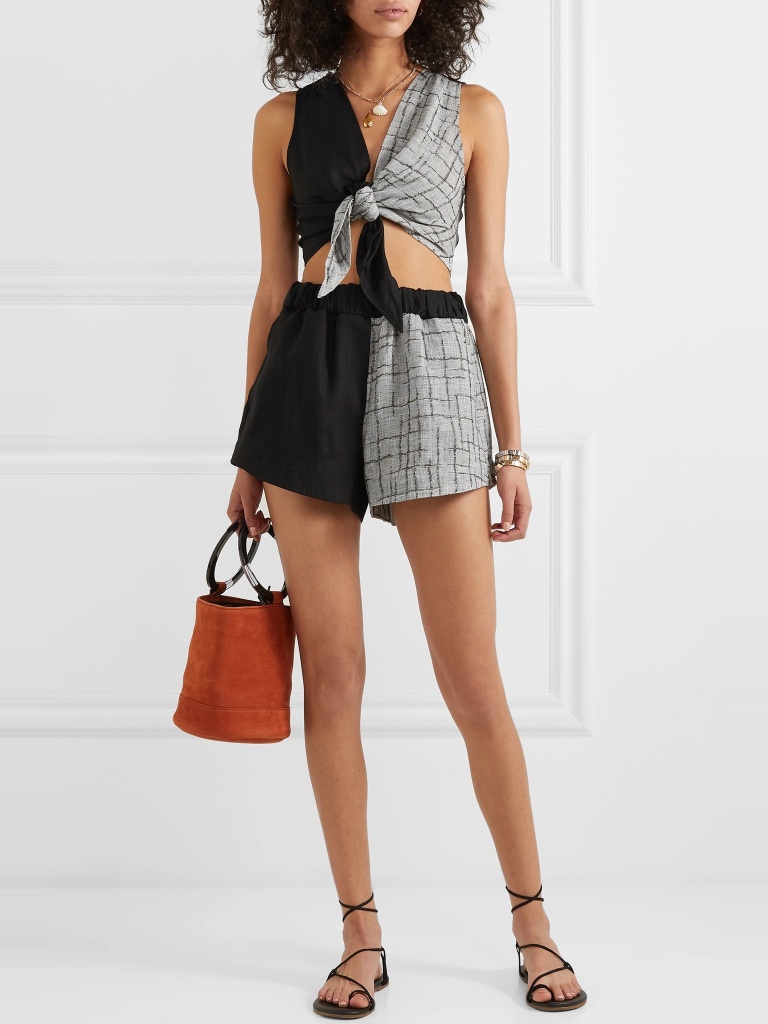}
\end{minipage}
\hspace{0.01cm}
\begin{minipage}[t]{0.18\linewidth}
\centering
\footnotesize{\textbf{Retrieved}}\\
\footnotesize{\textbf{(top-1)}}

\vspace{0.1cm}
\includegraphics[width=1.\linewidth]{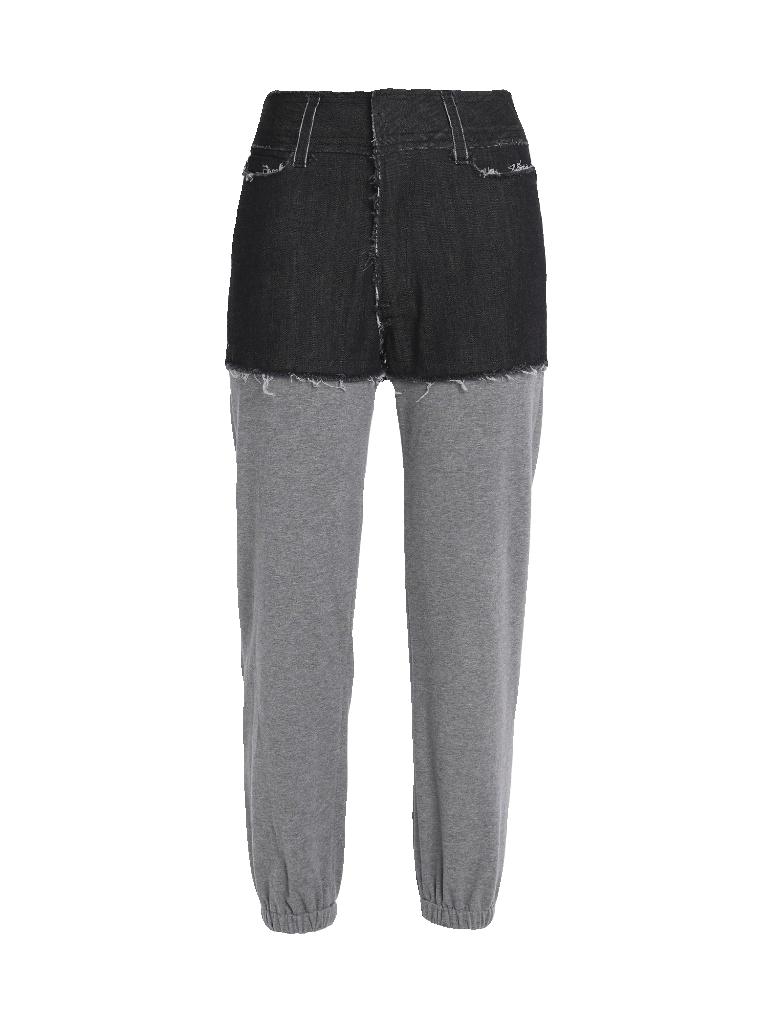}
\end{minipage}
\hspace{0.01cm}
\begin{minipage}[t]{0.18\linewidth}
\centering
\footnotesize{\textbf{Retrieved}}\\
\footnotesize{\textbf{(top-2)}}

\vspace{0.1cm}
\includegraphics[width=1.\linewidth]{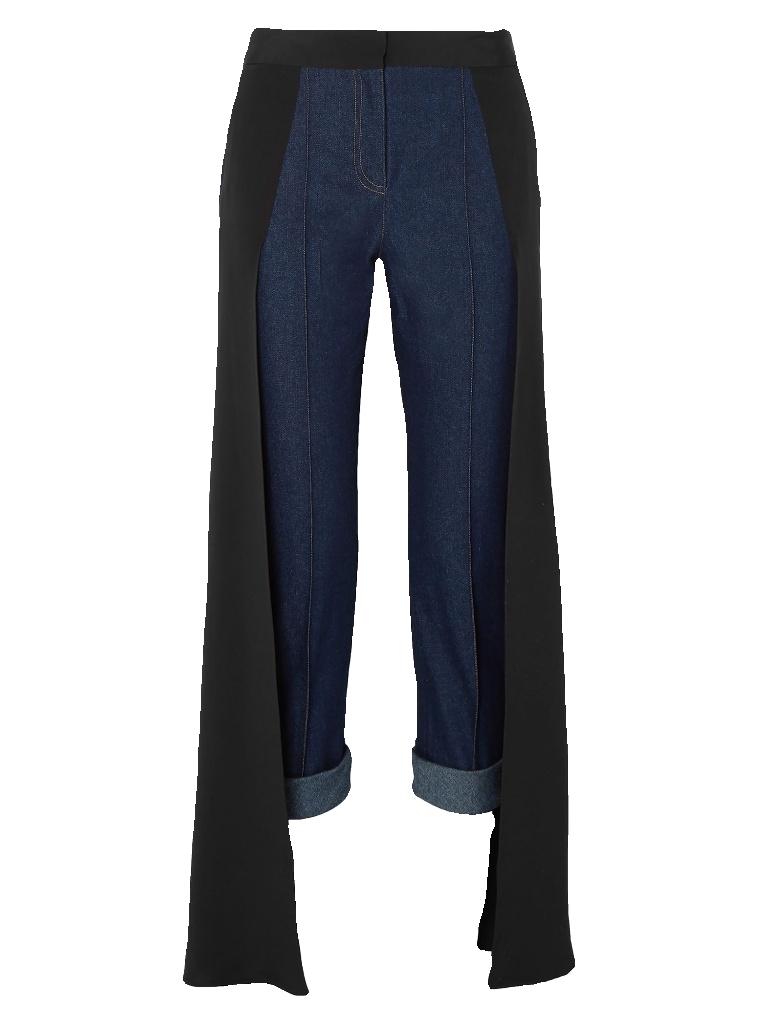}
\end{minipage}
\hspace{0.01cm}
\begin{minipage}[t]{0.18\linewidth}
\centering
\footnotesize{\textbf{Retrieved}}\\
\footnotesize{\textbf{(top-3)}}

\vspace{0.1cm}
\includegraphics[width=1.\linewidth]{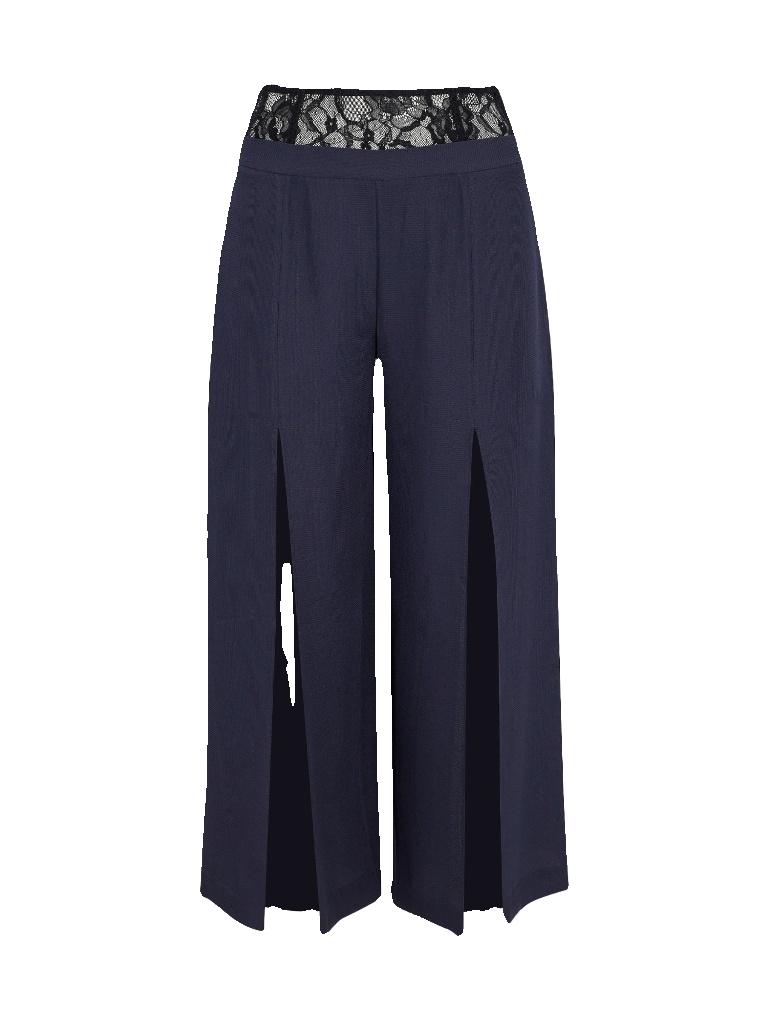}
\end{minipage}
\hspace{0.01cm}
\begin{minipage}[t]{0.18\linewidth}
\centering
\footnotesize{\textbf{w/ RAG}}\\
\footnotesize{($N_r=3$)}

\vspace{0.1cm}
\includegraphics[width=1.\linewidth]{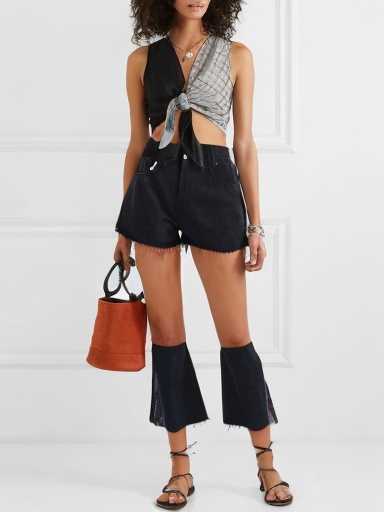}
\end{minipage}

\vspace{0.05cm}
\begin{minipage}[t]{0.18\linewidth}
\includegraphics[width=1.\linewidth]{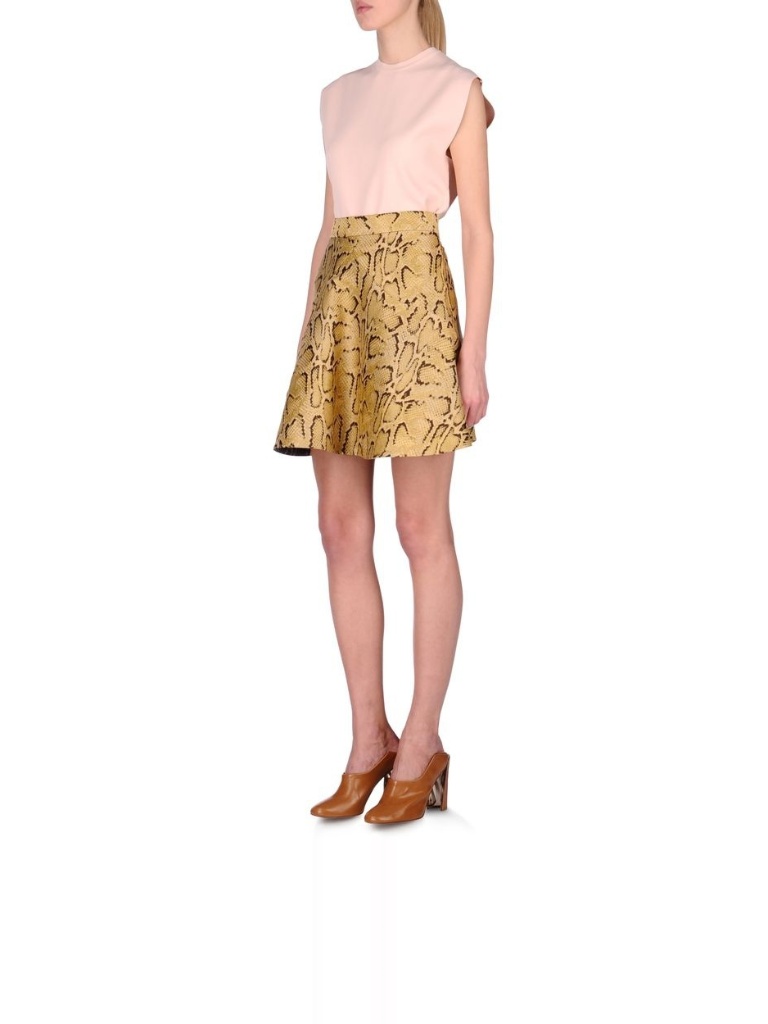}
\end{minipage}
\hspace{0.01cm}
\begin{minipage}[t]{0.18\linewidth}

\includegraphics[width=1.\linewidth]{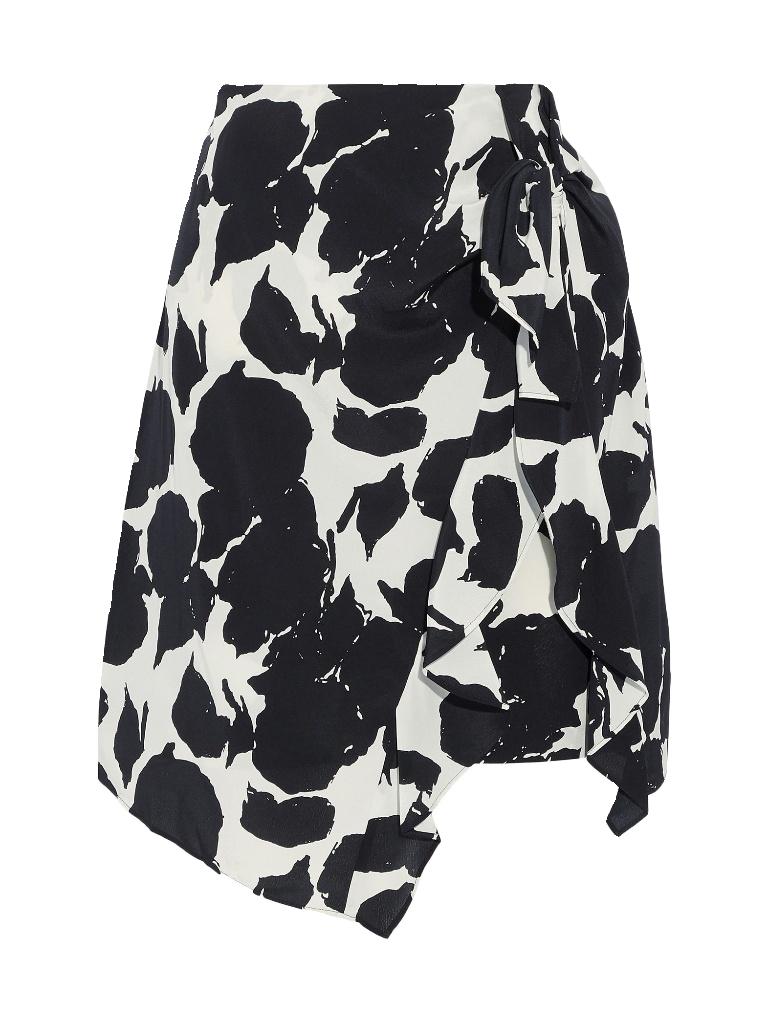}
\end{minipage}
\hspace{0.01cm}
\begin{minipage}[t]{0.18\linewidth}
\includegraphics[width=1.\linewidth]{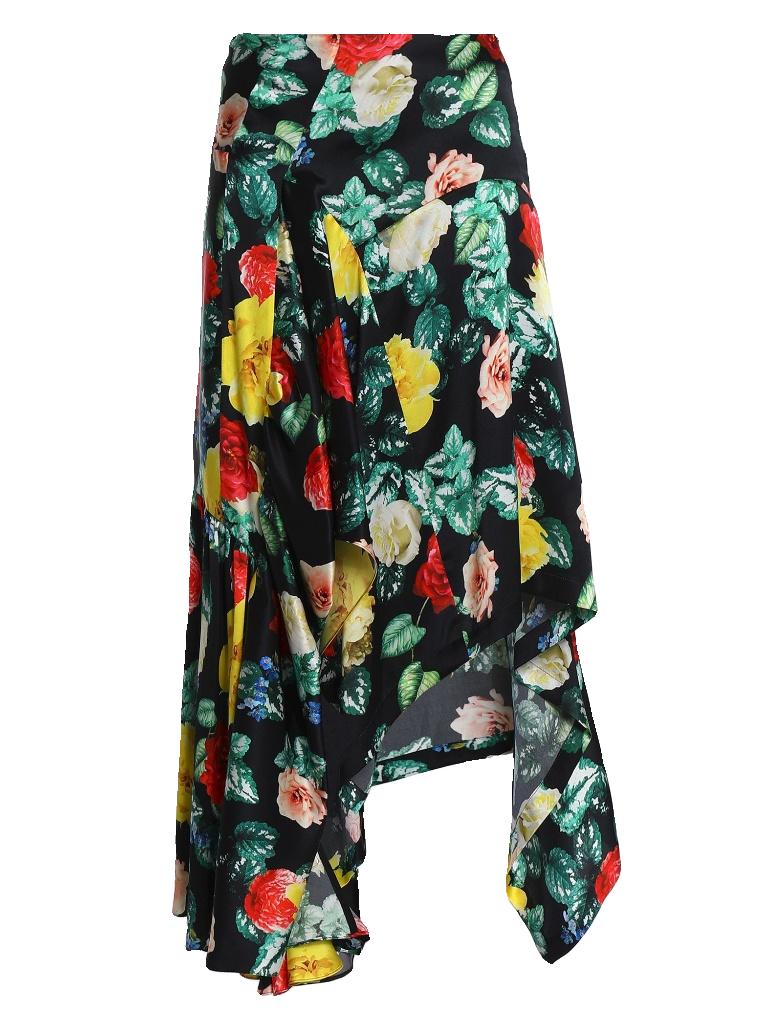}
\end{minipage}
\hspace{0.01cm}
\begin{minipage}[t]{0.18\linewidth}
\includegraphics[width=1.\linewidth]{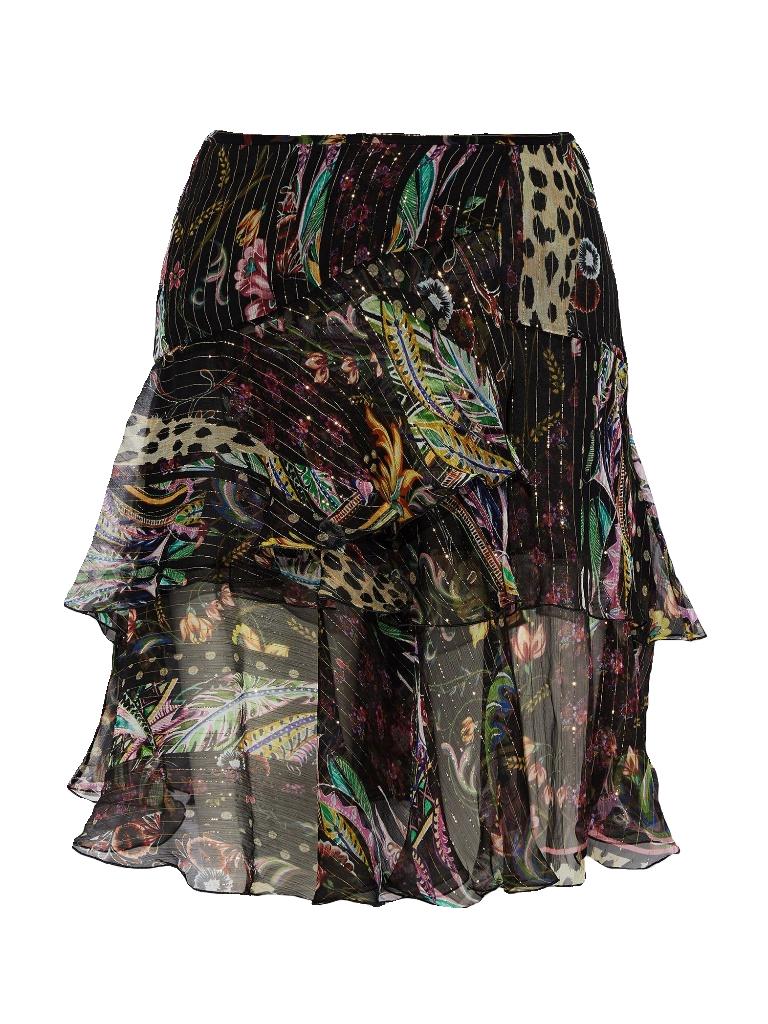}
\end{minipage}
\hspace{0.01cm}
\begin{minipage}[t]{0.18\linewidth}
\includegraphics[width=1.\linewidth]{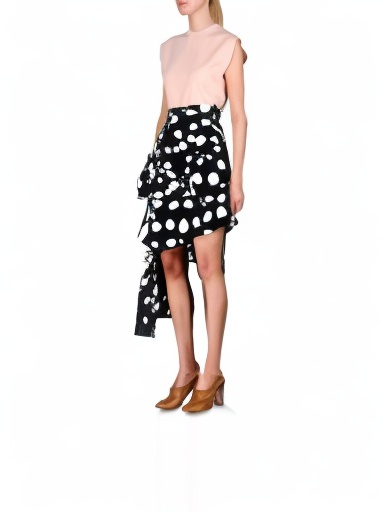}
\end{minipage}

\vspace{0.05cm}
\begin{minipage}[t]{0.18\linewidth}
\includegraphics[width=1.\linewidth]{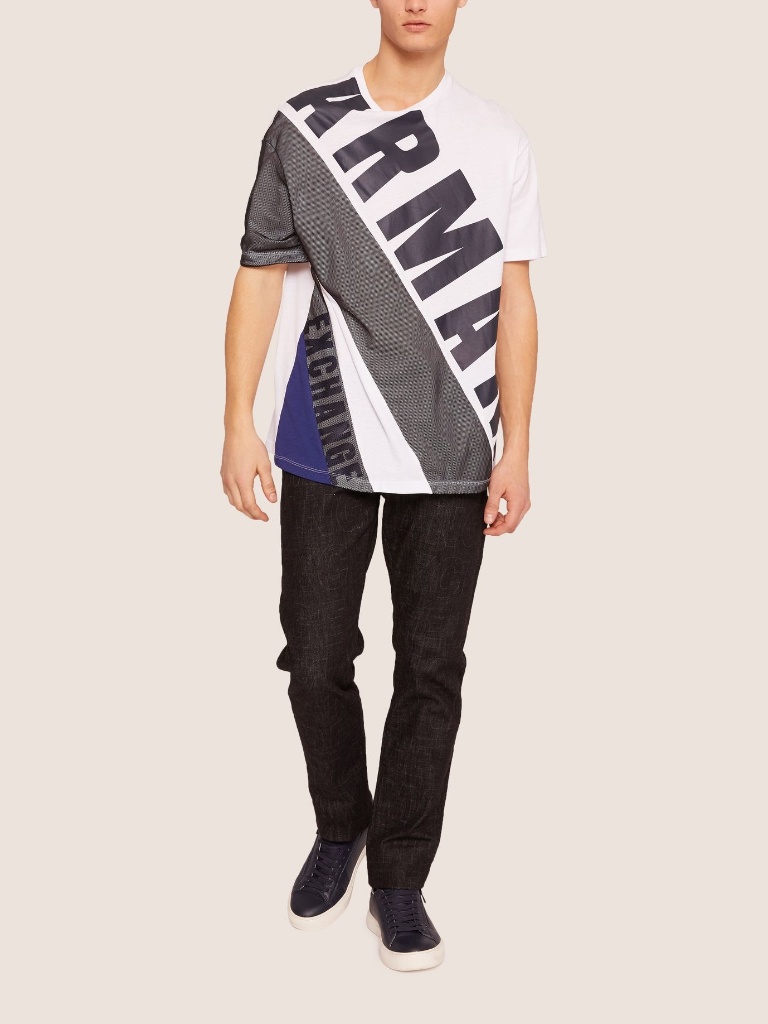}
\end{minipage}
\hspace{0.01cm}
\begin{minipage}[t]{0.18\linewidth}

\includegraphics[width=1.\linewidth]{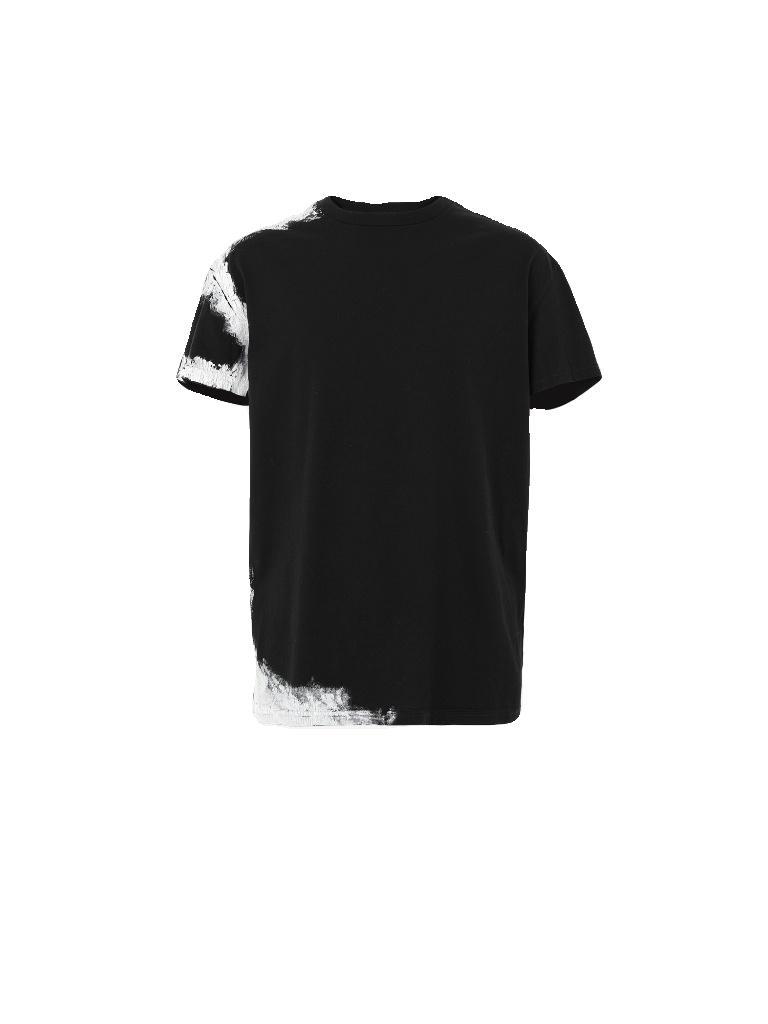}
\end{minipage}
\hspace{0.01cm}
\begin{minipage}[t]{0.18\linewidth}
\includegraphics[width=1.\linewidth]{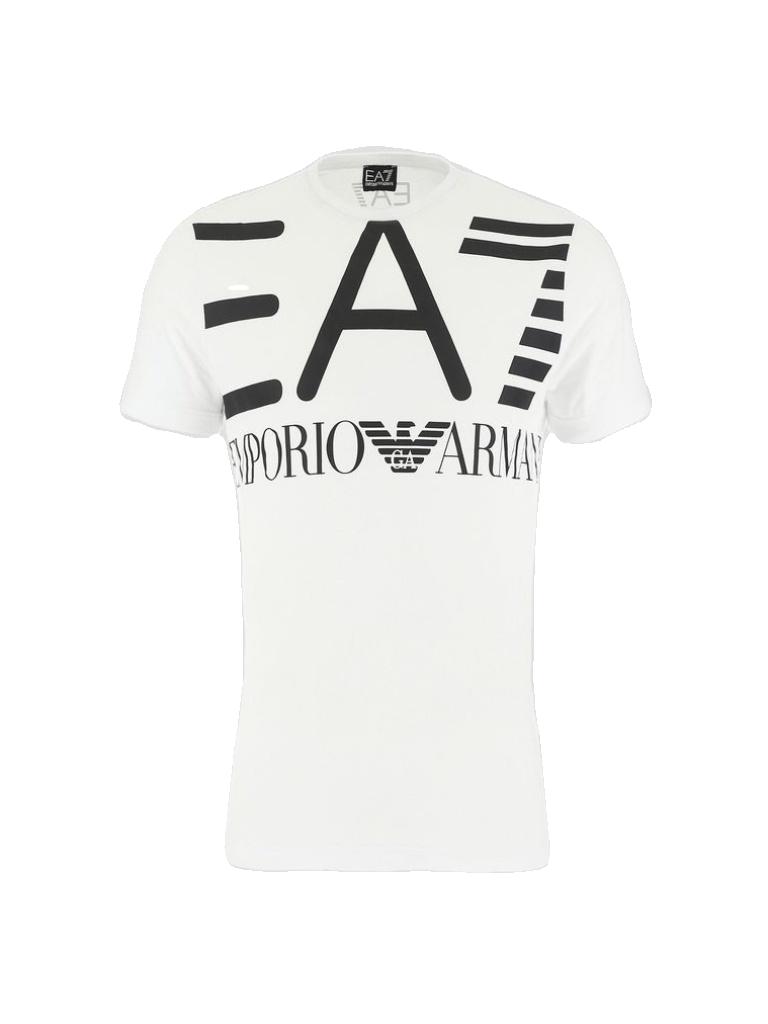}
\end{minipage}
\hspace{0.01cm}
\begin{minipage}[t]{0.18\linewidth}
\includegraphics[width=1.\linewidth]{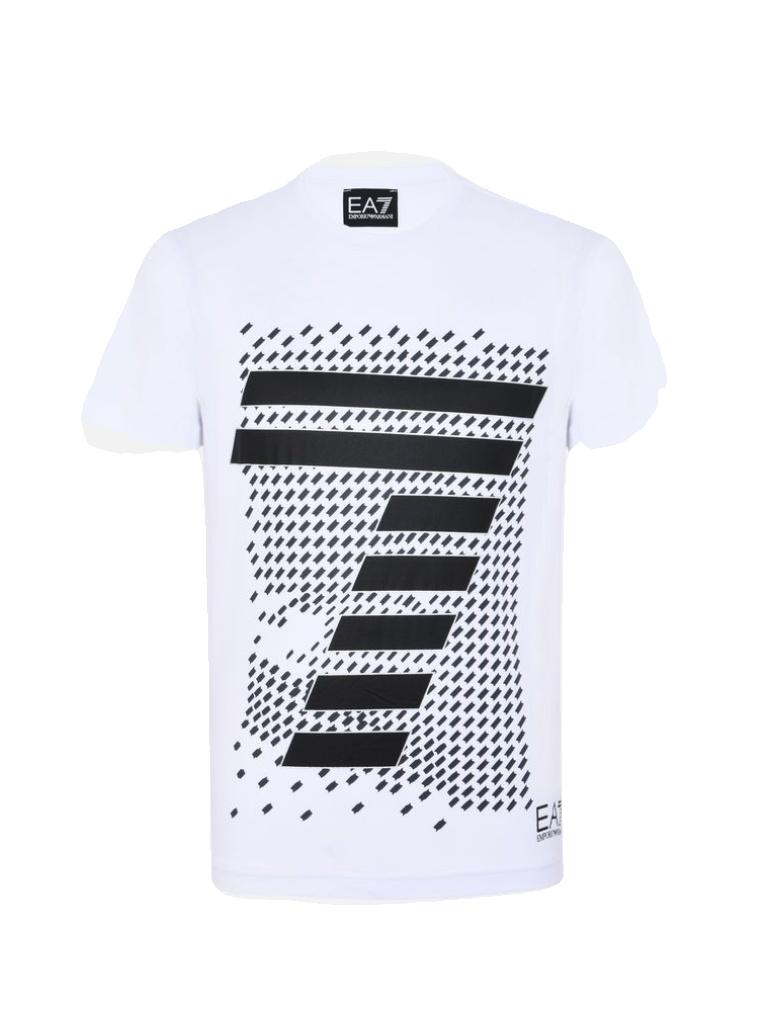}
\end{minipage}
\hspace{0.01cm}
\begin{minipage}[t]{0.18\linewidth}
\includegraphics[width=1.\linewidth]{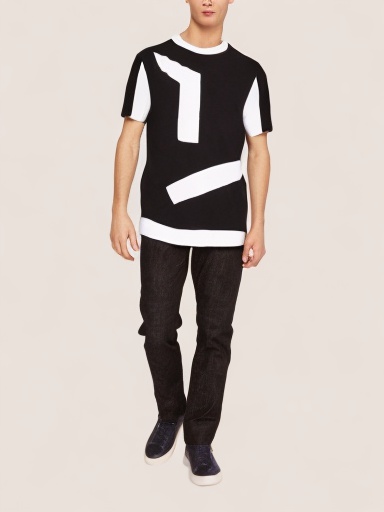}
\end{minipage}

\vspace{-0.25cm}
\caption{Sample failure case results showing the limitations of retrieval-augmented generation.}
\label{fig:fail}
\vspace{-0.3cm}
\end{figure}

\tit{Failure Cases} 
Finally, we present some failure cases in Fig.~\ref{fig:fail}. As observed, when the retrieved garments differ significantly in visual details (\eg, first and second rows), our model struggles to effectively combine the visual information from all retrieved items. Additionally, when the retrieved garments feature logos or text (\eg, third row), the generated image may not fully preserve the original characteristics. However, the model still transfers key visual features, such as colors and textures, from all retrieved elements.

\section{Conclusion}
\label{sec:conclusion}
In this work, we introduce \ours, the first retrieval-based framework for multimodal fashion image generation. This work opens a promising new avenue for the fashion industry by leveraging external image sources to enhance generative processes. Unlike traditional virtual try-on methods that require a specific garment input, our approach integrates a retrieval pipeline that effectively translates user-provided textual descriptions into high-quality, contextually relevant fashion images. Through extensive experiments, we validate the superior performance of \ours over competing methods, highlighting the effectiveness of retrieval in improving both the realism and coherence of generated images.

\section*{Acknowledgment}
We acknowledge the CINECA award under the ISCRA initiative, for the availability of high-performance computing resources. This work has been supported by the EU Horizon project ``ELIAS'' (No. 101120237).

\bibliographystyle{IEEEtran}
\bibliography{bibliography}

\end{document}